\documentclass[lettersize,journal]{IEEEtran}  

\usepackage{graphics} 
\usepackage{epsfig} 
\usepackage{times} 
\usepackage{amsmath} 
\usepackage{amssymb}  
\usepackage{arydshln}
\usepackage{multirow}

\usepackage{xcolor}
\usepackage{soul}
\usepackage{fixltx2e}

\usepackage[nobiblatex]{xurl}

\usepackage{color,soul}

\usepackage{hyperref}

\newcommand{\revise}[1]{\textcolor{black}{#1}}

\title{\LARGE \bf
Efficient Data-driven Joint-level Calibration of Cable-driven Surgical Robots
}

\author{Haonan~Peng, 
Andrew~Lewis,
Yun-Hsuan~Su, 
Shan~Lin,  
Dun-Tin~Chiang,
Wenfan~Jiang,
Helen~Lai,
Blake~Hannaford \\[1ex]
\thanks{Haonan Peng, Andrew Lewis, Tin Chiang, Helen Lai, and Blake Hannaford are with the University of Washington, Seattle, WA 98195, USA.
        {\tt\small {penghn, alewi, duntic, helenyl, blake}@uw.edu}}%
\thanks{Yun-Hsuan Su is with Mount Holyoke College, South Hadley, MA 01075, USA.
        {\tt\small msu@mtholyoke.edu}}%
\thanks{Shan Lin is with the University of California San Diego, 9500 Gilman Dr, La Jolla, CA 92093, USA.
        {\tt\small {shl102}@ucsd.edu}}%
\thanks{Wenfan Jiang is with the University of Michigan, Ann Arbor, MI 48109, USA.
        {\tt\small jiangwf@umich.edu}}%
\thanks{This paper is currently under review of npj Robotics.}%
}

\begin{document}
\bstctlcite{IEEEexample:BSTcontrol}

\maketitle
\thispagestyle{empty}
\pagestyle{empty}

\begin{abstract}

Knowing accurate joint positions is crucial for safe and precise control of laparoscopic surgical robots, especially for the automation of surgical sub-tasks. These robots have often been designed with cable-driven arms and tools because cables allow for larger motors to be placed at the base of the robot, further from the operating area where space is at a premium. However, by connecting the joint to its motor with a cable, any stretch in the cable can lead to errors in kinematic estimation from encoders at the motor, which can result in difficulties for accurate control of the surgical tool. 
In this work, we propose an efficient data-driven calibration of positioning joints of such robots, in this case the RAVEN-II surgical robotics research platform. While the calibration takes only 8--21 minutes, the accuracy of the calibrated joints remains high during a 6-hour heavily loaded operation, suggesting desirable feasibility in real practice. The calibration models take original robot states as input and are trained using zig-zag trajectories within a desired sparsity, requiring no additional sensors after training. Compared to fixed offset compensation, the Deep Neural Network (DNN) calibration model can further reduce $76\%$ of error and achieve accuracy of $0.104^{\circ}$, $0.120^{\circ}$, and $0.118$ mm in joints 1, 2, and 3, respectively. In contrast to end-to-end models, experiments suggest that the DNN model achieves better accuracy and faster convergence when outputting the error to correct original inaccurate joint positions. Furthermore, a linear regression model is shown to have 160 times faster inference speed than DNN models for application within the RAVEN's 1000 Hz servo control loop, with slightly compromised accuracy. Application of this study's methodology should significantly improve the accuracy and repeatability of similar cable-driven robots without access to direct joint position data.


\end{abstract}

\section{Introduction}
With help from Robotic Surgical Assistants (RSAs) such as the da Vinci Robotic Surgical Systems, Robot-assisted Minimally Invasive Surgery (RAMIS) improves patient outcomes and has seen accelerating implementation for over 20 years \cite{sayari2019review, peters2018review, chughtai2015national, haidegger2022robot}. Benefiting from cable-driven joints, RSA research platforms such as RAVEN-II \cite{hannaford2012raven} and the da Vinci Research Kit (dVRK) \cite{kazanzides2014open} have lightweight and compact arms and tools \cite{rosen2011surgical, palep2009robotic, ramadurai2012application}. However, sterilization \cite{josephs2021medical} and the complexity of wiring make it difficult and costly to implement encoders directly on the joints of RAVEN-II \cite{kosari2013control}. Joint positions obtained from motor encoders at the robot base can have considerable transmission errors and undermine the estimation of the end-effector pose derived by forward kinematics \cite{peng2020real, haghighipanah2015improving}. Although in bilateral leader and follower control \cite{mehrdad2020review}, human operators can still perform accurate operations with visual feedback \cite{kim2015effects, maza2020past}, however, endoscope-based visual tracking in practice can be challenging because of dynamic and reflective surgical scenes \cite{lin2016video, lin2020lc}. As a consequence, inaccurate robot pose estimation makes the safe automation of surgical sub-tasks in clinical settings challenging, such as automatic ablation\cite{hu2015semi}, cutting \cite{thananjeyan2017multilateral}, palpation \cite{garg2016tumor}, suturing \cite{pedram2017autonomous, jiang2023markerless, varier2020collaborative}, debridement \cite{kehoe2014autonomous, seita2018fast}, manipulation of tissue \cite{wang2018unified, alambeigi2018toward}, and peg transfer \cite{xu2021surrol, rahman2021sartres, gonzalez2021deserts, hwang2022automating}. For these tasks, accurate pose estimation is crucial and is generally achieved through painstaking calibration \cite{kehoe2014autonomous} or with visual tracking, which may not be feasible in clinical applications. Moreover, for intelligent surgical robot agents for improved surgeon-robot interaction, such as tremor canceling \cite{riviere2003toward}, motion compensation of the dynamic surgical scenes \cite{lindgren2017towards, yuen2009robotic}, and vision-based force estimation \cite{su2018real}, precise end-effector pose of the surgical robot is also required \cite{li2019raven}.

To alleviate the inaccuracy of cable-driven surgical robots, both model-based and data-driven methods show desirable performance. Model-based methods benefit from expert analysis of the robot design and have better explainability in general. However, accurate modeling of cable-driven mechanisms can be challenging with consideration of practical cable effects \cite{ramadurai2012application, naerum2009robustness, Choi2017Tension}. To model cable effects, Miyasaka et al. \cite{miyasaka2016hysteresis} utilize a Bouc-Wen hysteresis model and a linear damper to model longitudinally loaded cables, in which 9 hysteresis model parameters are optimized by a genetic algorithm. Haghighipanah et al. \cite{haghighipanah2015improving} apply an unscented Kalman Filter (UKF) to obtain a more accurate estimation of the positioning joints of RAVEN-II. A significant improvement in accuracy can be found in joints 2 ($1.669^{\circ}$) and joint 3 ($0.928$ mm), however, joint 1 ($1.295^{\circ}$) sees less reduced error due to relatively higher joint stiffness.

On the other hand, by utilizing features from robot states, data-driven (preferably learning-based) methods can achieve state-of-the-art accuracy with sufficient training data \cite{hwang2020efficiently, mahler2014learning}. With the effectiveness of data-driven methods proved by many, the efficiency of calibration time, training data, and real-time inference remains less investigated. Huwang et al. present a calibration based on a recurrent neural network that reduced the estimation error of the dVRK end-effector position from 2.96 mm to 0.65 mm \cite{hwang2020efficiently}. This calibration \revise{takes 31 minutes and} further helps automate surgical peg transfer and achieve superhuman performance \cite{hwang2022automating}. Seita et al. present a two-phase calibration, a DNN model for coarse calibration and then a random forest model for fine training \cite{seita2018fast}. Mahler et al. develop a Gaussian process regression calibration for RAVEN-II, in which velocity is found to be an important input for the model \cite{mahler2014learning}. To ensure the accuracy of supervised data-driven calibration, reliable ground truth is of importance. Visual-based tracking shows considerable feasibility and accuracy. In \cite{peng2020real} and \cite{hwang2020efficiently}, ground truth is obtained by tracking colored spheres mounted on the robot end-effector using 4 monocular cameras or 1 RGB-D camera. In \cite{mahler2014learning}, ground truth is obtained by tracking LEDs mounted on the end-effector.

In this paper, an efficient data-driven joint calibration of RAVEN-II is proposed. Zig-zag calibration trajectories are used for desirable coverage of the workspace. By studying the effectiveness of the direction and sparsities of the calibration trajectories, the calibration can be finished in 8 to 21 minutes and has an accuracy of $0.104^{\circ}$, $0.120^{\circ}$, and $0.118$ mm in joints 1, 2, and 3, respectively, reducing more than $76\%$ of the error. Although continuous operating and load cause decreased accuracy, with sufficient training data (maximum of 17 minutes) the accuracy remains within $0.855^{\circ}$, $0.432^{\circ}$, and $0.181$ mm for the positioning joints in 6-hour heavily loaded operating. To achieve robust calibration performance, the choice of input-output of the calibration models, and the robot homing procedure are also studied.

This work makes the following contributions: 1) an efficient joint calibration pipeline that costs less than 21 minutes and improves the accuracy significantly; 2) evaluation of calibration performance on 6-hour idleness, unloaded, and loaded operating, in the time scale of real surgeries; 3) besides the DNN calibration with best accuracy, a linear regression calibration established with slightly compromised accuracy but fast enough inference speed to meet the robot's 1000 Hz servo control; 4) a general software controller for RAVEN-II robot, to the best of the authors' knowledge, is the only one providing a Python API.
\section{Result} \label{char_result}
\subsection{Calibration Workflow} \label{schar_Workflow}

\begin{figure*}
\centering
\vspace{0.3em}
\includegraphics[width=0.9\textwidth]{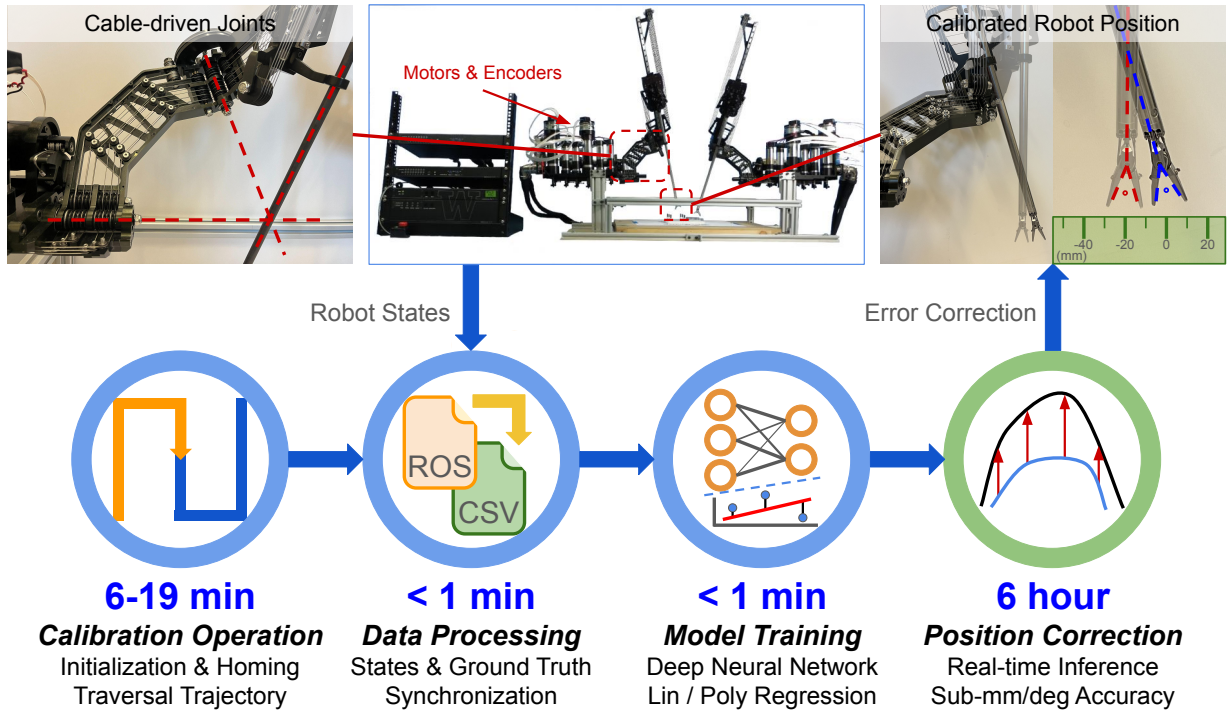}
\vspace{-1.em}
\caption{\revise{Workflow of the calibration. The time cost mainly depends on the collection of training data. The time cost of data processing and model training may vary for different CPUs. GPU is not necessary for the training of neural network models in this paper, unless stated otherwise. Designed for surgical purposes, the first 3 joint axes of RAVEN-II always intersect at the same location which is called the remote motion center and the surgical instrument will always pass this location as the incision site (top left). In order to possess compact and light arms, the motors and encoders are mounted on the robot base, and joints are driven by cables, which results in considerable errors in joint positions obtained by remote motor encoders (top middle). Without calibration, the average positional error of RAVEN-II's end-effector is around 20 mm (top right).}} 
\vspace{-1.5em}
\label{fig_workflow}
\end{figure*}

The objective of the proposed data-driven calibration is to achieve better accuracy of the positioning joints of the RAVEN-II surgical robot in an efficient manner. As shown in Fig. \ref{fig_workflow}, the calibration workflow contains the following sequential procedures:

\textbf{Calibration operation:} Firstly, it takes less than 2 minutes to initialize the robot and perform the homing procedure, during which the robot explores joint limits and registers the motor encoders. If calibration is to be performed, temporary encoders for ground truth collection are mounted on the joints and are registered to joint positions at the same time. 
Then, to collect training data, the robot follows predefined zig-zag trajectories which have even distribution in the workspace of the robot (details in \ref{sch_wkspc_n_traj} and \ref{sch_crtk_controller}). Meanwhile, the robot states and ground truth joint positions are recorded as ROS bags \cite{quigley2009ros}. Depending on the sparsity and the size of the training dataset, this procedure takes 4-17 minutes. Longer trajectories with smaller sparsity have better coverage of the robot workspace, but also require a longer time to record.

\textbf{Data processing:} This step extracts data from the recorded ROS bags from the previous step and saves them into .csv files. Synchronization between robot states and ground truth joint positions is also used to construct training pairs of input robot state and output joint positions. This step takes less than 1 minute with the largest dataset in this paper.

\textbf{Model training:} With the dataset from the previous step, calibration models can be trained/fitted. Deep neural network (DNN), linear regression, and 2nd-order polynomial regression models are studied in this paper (details in \ref{sch_dnn_reg_models}). The models take robot states as input and output calibrated joint positions. With optimized input dimension and DNN structure, all models can be trained within 1 minute (without GPU for the DNN model).

\textbf{Position correction:} After training, the calibration models are able to work in parallel with the robot control system in real time. The models take the original robot states as input so that no additional sensor is needed. The output is the corrected joint positions with sub-mm/deg accuracy and the accuracy remains desirable in 6-hour continuous loaded and unloaded operating.    


\subsection{Robot Workspace and Calibration Trajectories} \label{sch_wkspc_n_traj}

\begin{figure*}
\centering
\includegraphics[width=0.8\textwidth]{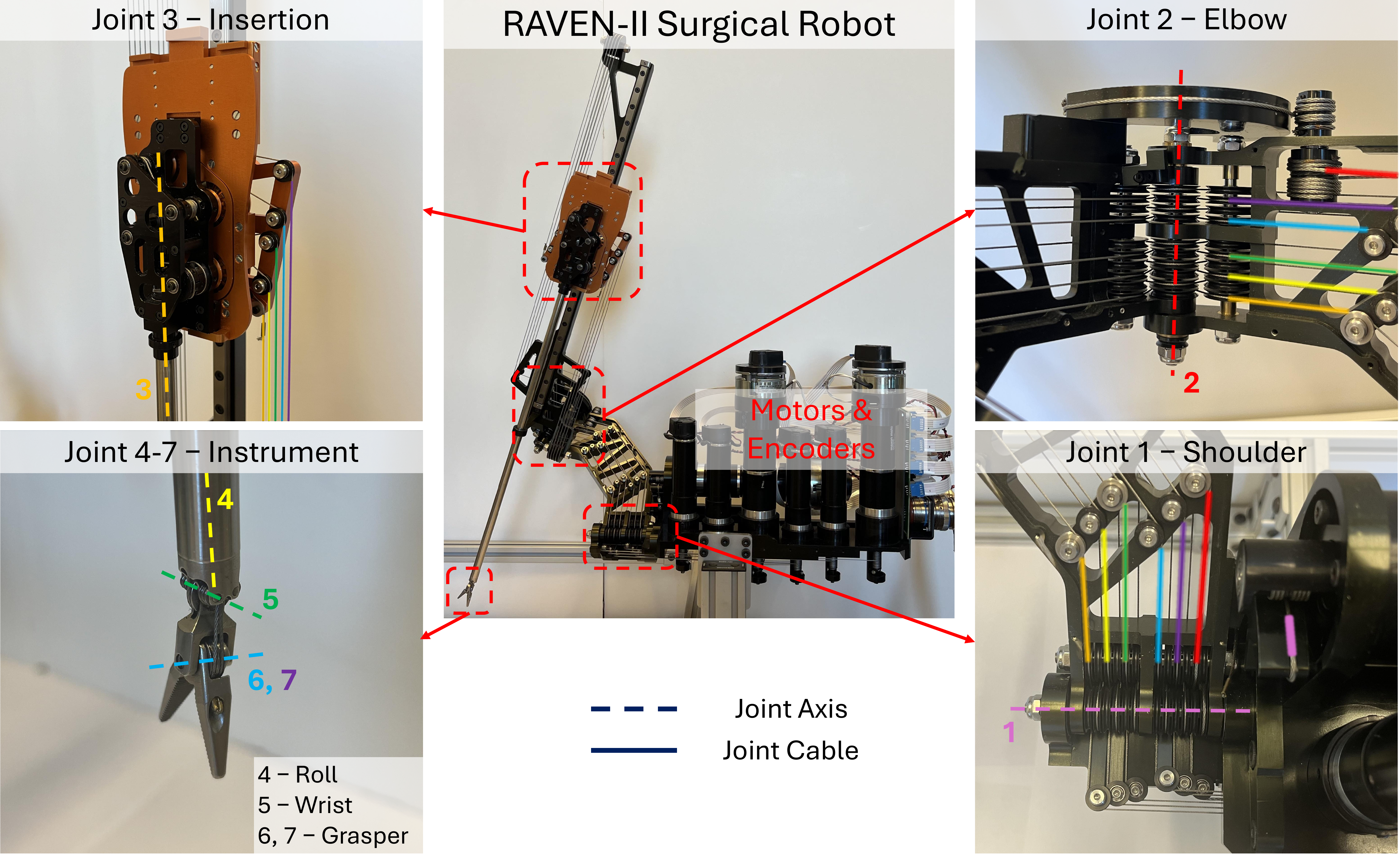}
\caption{The joints and cable connections of the RAVEN-II surgical robot. Compared to joints 2-7, joint 1 has a considerably short cable length (alpha wrap). Joint 3 is prismatic and has a large range for insertion, while all the rest of the joints are rotational. Joints 6 and 7 control the left and right fingers of the end-effector and can be considered as one joint in kinematics. Thus, RAVEN-II has 6 joints in kinematic models instead of 7. The left arm and the right arm of RAVEN-II are symmetric.} 
\label{fig_raven_cable}
\end{figure*}

This paper studies the positioning joints (joints 1-3) of the RAVEN-II robot. Fig. \ref{fig_raven_cable} shows the joints and cable connections of RAVEN-II. Designed for RAMIS, the last 4 joints of RAVEN-II are concentrated near the end-effector with very small (13 mm) or zero link lengths, and thus have much less effect on the location of the end-effector, compared to the first 3 joints. 

\begin{figure*}
\centering
\vspace{0.3em}
\includegraphics[width=0.8\textwidth]{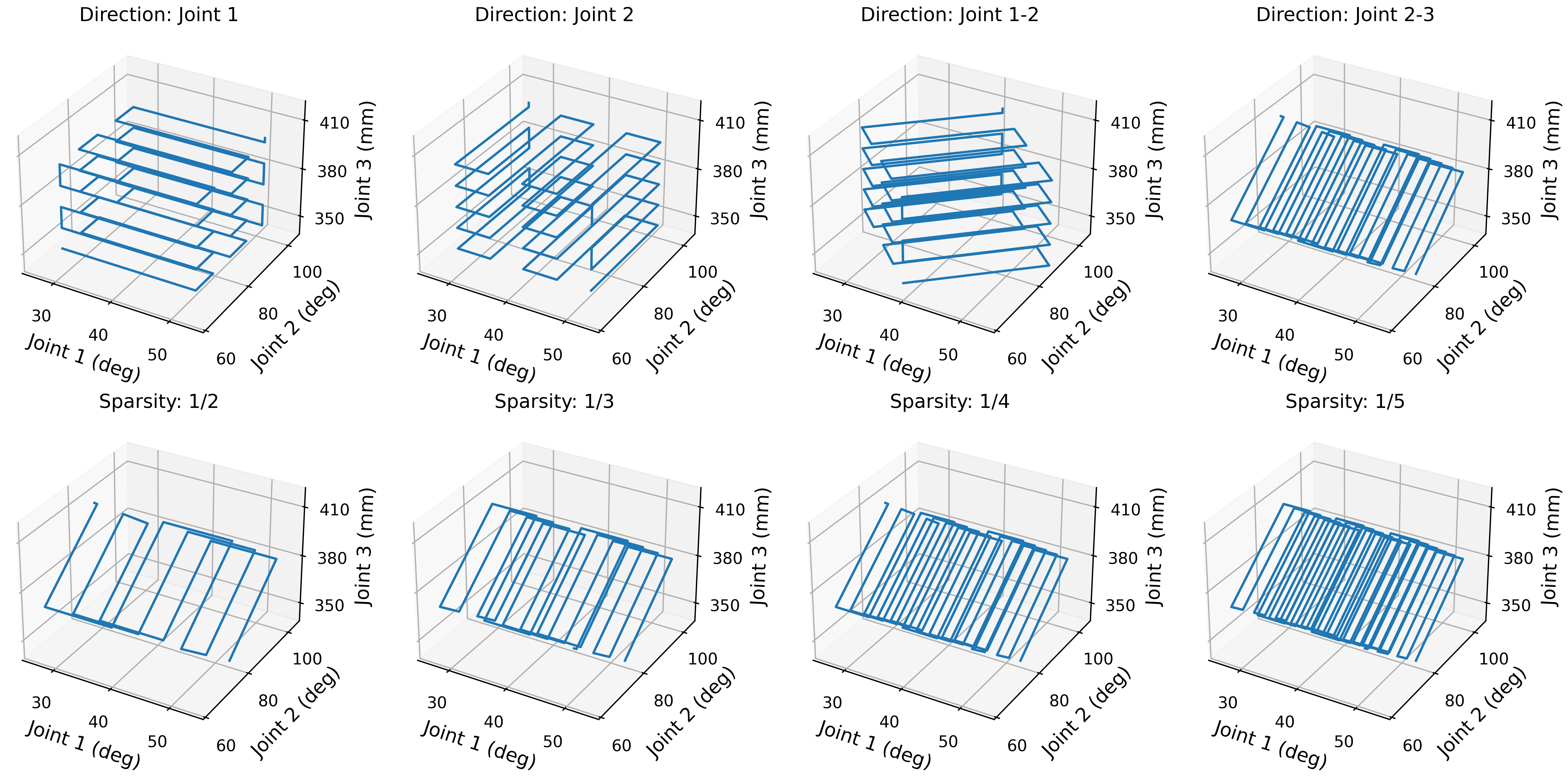}
\vspace{-1.em}
\caption{Examples of calibration trajectories with different directions (top) and sparsity (bottom). Please note that the trajectories are defined in joint space instead of Cartesian space. Different directions cause different coverage in different joints. Smaller sparsity results in better coverage of the joint space, but also takes a longer time to execute.} 
\vspace{-1.5em}
\label{fig_diff_trajs}
\end{figure*}

Zig-zag trajectories with different directions and sparsity are chosen as calibration trajectories for data collection, as Fig. \ref{fig_diff_trajs} shows. \revise{With limited calibration time, zig-zag trajectories ensure even coverage of the robot workspace and prevent undiscovered areas.} However, because trajectories are continuous, different directions of the trajectories have different coverage in different joints. For example, as the top left sub-plot in Fig. \ref{fig_diff_trajs} shows, the zig-zag trajectory has a sparsity in the direction of joints 2 and 3, but has no sparsity in the direction of joint 1. The bottom row of Fig. \ref{fig_diff_trajs} shows different sparsity, smaller sparsity gives better coverage of the workspace but requires a longer time for the calibration. 

The calibration zig-zag trajectories are defined as a set of sequential 3D vectors representing points in the workspace of the positional joints of the robot. \revise{All trajectories are obtained by rotating and translating $\mathbb{T}_1$ -- a zig-zag trajectory with the direction of joint 1, and are first defined on a normalized range $(-0.5,  0.5)$ and will be scaled according to each joint's limit.} For example, the trajectory with the direction of joint 2 is obtained by
\begin{equation}
\begin{aligned}
\mathbb{T}_2 = T(\begin{bmatrix}0.5\\0.5\\0.5\end{bmatrix})R(J_3, 90^{\circ})\mathbb{T}_1 
\label{equ_traj_1}
\end{aligned}
\end{equation}
where $T()$ and $R()$ are operators of translation and rotation, respectively, $J_3$ means the axis of joint 3 in the joint space, the trajectory with the direction of joint 1 -- $\mathbb{T}_1$ is a $4 \times N$ matrix with the direction of joint 1, in which the first 3 rows are positions of joint 1, 2, and 3, respectively. In our implementation, both translation and rotation are applied by homogeneous transformation matrices.

Similarly, trajectories with the direction of joint 1-2 and joint 1-2-3 are generated by the following equations, respectively.
\begin{equation}
\begin{aligned}
\mathbb{T}_{1-2} = T(\begin{bmatrix}0.5\sqrt{2}\\0.5\sqrt{2}\\0.5\end{bmatrix})R(J_3, 45^{\circ})\mathbb{T}_1 \odot \begin{bmatrix}1/\sqrt{2}\\1/\sqrt{2}\\1\end{bmatrix}
\label{equ_traj_2}
\end{aligned}
\end{equation}

\begin{equation}
\begin{aligned}
\mathbb{T}_{1-2-3} = T(\begin{bmatrix}0.5\sqrt{3}\\0.5\sqrt{3}\\0.5\sqrt{3}\end{bmatrix})R(J_2, 45^{\circ})R(J_1, 45^{\circ})\mathbb{T}_1 \odot \begin{bmatrix}1/\sqrt{3}\\1/\sqrt{3}\\1/\sqrt{3}\end{bmatrix}
\label{equ_traj_3}
\end{aligned}
\end{equation}
where $\odot$ indicates elemental-wise multiplication. After Equation (\ref{equ_traj_1})-(\ref{equ_traj_3}), joint limit of all trajectories are within $(0, 1)$, though with different direction.

In this paper, the following directions of zig-zag trajectories are used: \textbf{1) single direction}: Joint 1, Joint 2, and Joint 3; \textbf{2) double direction}: Joint 1-2, Joint 2-3, Joint 1-3; \textbf{3) triple direction}: Joint 1-2-3. Given:
\begin{equation}
\begin{aligned}
c = 0.5(J_{max}+J_{min})
\label{equ_traj_4}
\end{aligned}
\end{equation}

\begin{equation}
\begin{aligned}
r = J_{max}-J_{min}
\label{equ_traj_5}
\end{aligned}
\end{equation}
where $c$ and $r$ are the center position and range of a joint, $J_{max}$ and $J_{min}$ are the maximum and minimum position of a joint. Then trajectories are translated and scaled so that a single direction has a range $(c-\frac{1}{2\sqrt{3}}r, c+\frac{1}{2\sqrt{3}}r)$, a double direction has a range $(c-\frac{\sqrt{2}}{2\sqrt{3}}r, c+\frac{\sqrt{2}}{2\sqrt{3}}r)$, and a triple direction has a range $(c-\frac{1}{2}r, c+\frac{1}{2}r)$. 
These different ranges result in all trajectories having the same center and covering the same volume in the normalized joint workspace.


\subsection{Robot States and Calibration Models} \label{sch_dnn_reg_models}
\begin{figure}
\centering
\vspace{0.5em}
\includegraphics[width=0.40\textwidth]{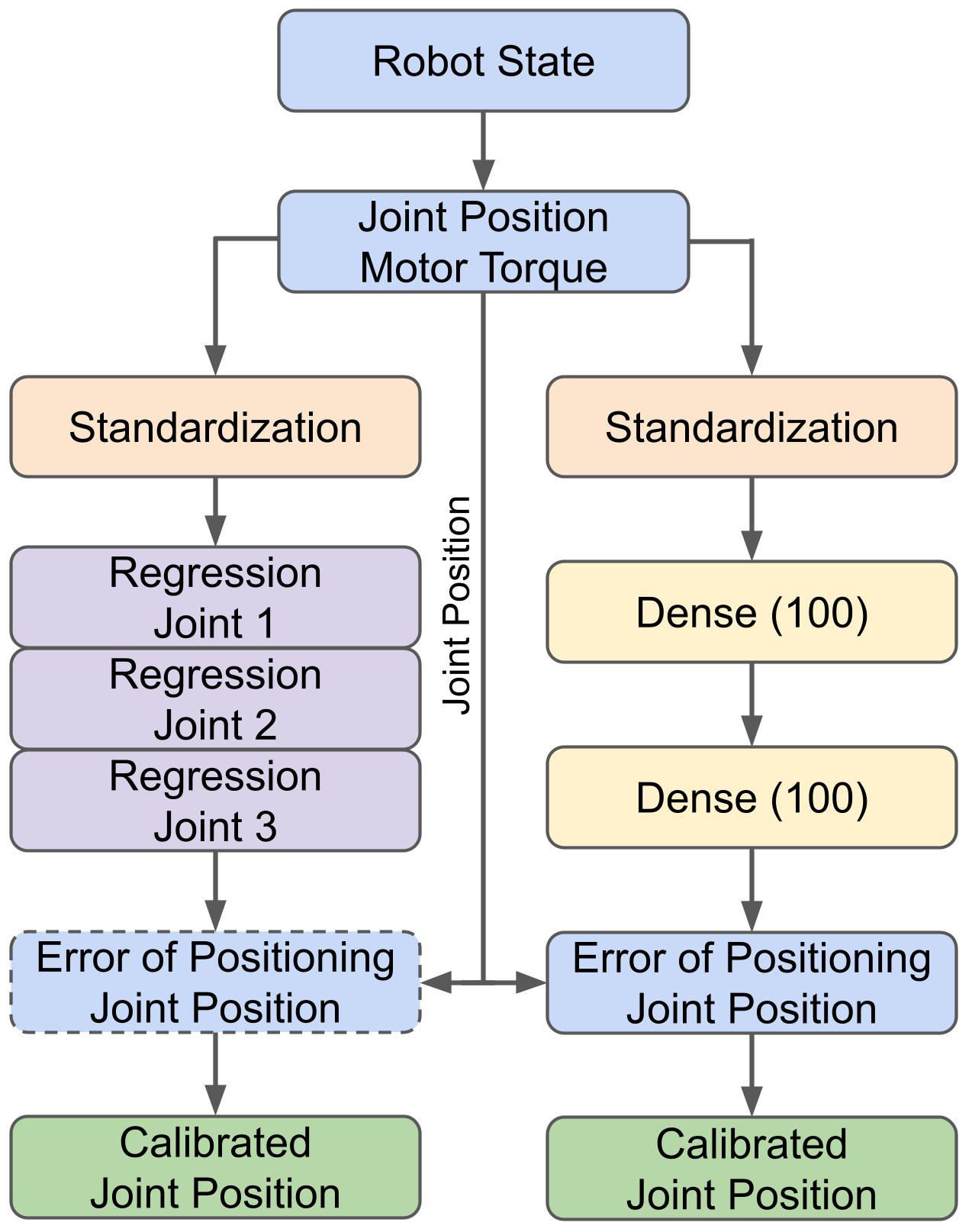}
\caption{Model of regression-based calibration (left) and learning-based calibration (right). The outputs of the DNN model and regression model are the errors of the joint positions, these errors are added back to RAVEN-II's original inaccurate joint positions to obtain the calibrated joint positions. The output of the regression models can be either the same or end-to-end output of joint positions (details in \ref{schar_train_on_err}).} 
\vspace{-1.5em}
\label{fig_learn_reg_model}
\end{figure}

The calibration models based on DNN and regressions are shown in Fig. \ref{fig_learn_reg_model}. The details about the hyperparameters of the DNN model can be found in \ref{schar_exp_setup_hyper_param}. The input of the models is obtained from the robot state 'ravenstate' which is a ROS topic published at 1000 Hz in real time. Based on the ablation study \cite{peng2023ablation}, the following features are used as input: \textbf{Current joint positions} are the built-in joint positions derived from motor positions, and the accuracy is degraded by the cable-driven mechanism. A typical arm of RAVEN-II has 7 joints. The positioning joints (joints 1-3) mainly determine the position of the end-effector, while the orientational joints (joints 4-7) mainly determine the orientation of the end-effector. \textbf{Motor torques} are derived from motor current control commands. RAVEN-II does not have any direct measurement of joint torques. More details about the robot states are in \ref{schar_robot_states}.

It is possible to include all features in the robot state as input for the calibration models. However, for linear and 2nd-order polynomial regression models, an increase in the input dimension can result in significantly more parameters. With limited training data, there can be a higher risk of overfitting, especially for 2nd-order polynomial regression. Choices of input features are studied in \ref{schar_robust_features}.

Instead of end-to-end calibration models that output calibrated joint positions, the output is the error of the original joint positions in the robot state. Then, the calibrated joint positions are obtained by using the output error to correct the original inaccurate joint positions. Training on errors allows better performance for the DNN models, while it is optional for regression models (details in \ref{schar_train_on_err}). 

Besides learning and regression calibration, fixed offset compensation is also applied as a baseline method. This calibration uses the same training data and computes the average bias between the robot's original joint positions and the ground truth joint positions. Then each joint position is compensated with the average bias when inferred.



\subsection{Datasets} \label{sch_datasets}
Data was collected as ROS bags (details in \ref{sch_crtk_controller}) and then processed to form training and test datasets. Details about the experiment setup can be found in \ref{schar_exp_setup_hyper_param},  The external joint positions were recorded under 100Hz, while the robot state 'ravenstate' was recorded under 30Hz due to high dimensions. Synchronization was performed when processing ROS bags. In the experiments, different training trajectories with different directions and sparsity were recorded and the details are included in the following sections, together with experiment results.

\subsection{Directions of Calibration Trajectories} \label{schar_exp_dir}

\begin{table*}[h]
\vspace{1.em}
\caption{\vspace{-0.3em} Learning and Regression Calibration Error with Different Directions of Trajectories}
\vspace{-1.5em}
\begin{center}
\begin{tabular}{ccccccccc}
\hline
\multicolumn{1}{c|}{\multirow{2}{*}{\textbf{Direction}}} & \multicolumn{2}{c|}{\textbf{DNN}}                              & \multicolumn{2}{c|}{\textbf{2nd Reg}}                          & \multicolumn{2}{c|}{\textbf{Lin Reg}}                          & \multicolumn{1}{c|}{\textbf{Fixed Offset}} & \textbf{Raw} \\
\multicolumn{1}{c|}{}                                    & RMSE           & \multicolumn{1}{c|}{Percentage}               & RMSE           & \multicolumn{1}{c|}{Percentage}               & RMSE           & \multicolumn{1}{c|}{Percentage}               & \multicolumn{1}{c|}{RMSE}              & RMSE         \\ \hline
\multicolumn{9}{c}{Joint 1 (deg)}                                                                                                                                                                                                                                                                                   \\ \hline
\multicolumn{1}{c|}{J1}                                  & 0.215          & \multicolumn{1}{c|}{\textit{25.0\%}}          & 0.193          & \multicolumn{1}{c|}{\textit{22.5\%}}          & 0.214          & \multicolumn{1}{c|}{\textit{25.0\%}}          & \multicolumn{1}{c|}{0.857}             & 1.930        \\
\multicolumn{1}{c|}{J2}                                  & 0.186          & \multicolumn{1}{c|}{\textit{20.4\%}}          & 0.156          & \multicolumn{1}{c|}{\textit{17.1\%}}          & 0.214          & \multicolumn{1}{c|}{\textit{23.5\%}}          & \multicolumn{1}{c|}{0.911}             & 2.015        \\
\multicolumn{1}{c|}{J3}                                  & \textbf{0.115} & \multicolumn{1}{c|}{\textit{\textbf{13.1\%}}} & 0.144          & \multicolumn{1}{c|}{\textit{16.5\%}}          & \textbf{0.198} & \multicolumn{1}{c|}{\textit{\textbf{22.6\%}}} & \multicolumn{1}{c|}{0.877}             & 2.061        \\
\multicolumn{1}{c|}{J1-J2}                               & 0.209          & \multicolumn{1}{c|}{\textit{23.1\%}}          & 0.169          & \multicolumn{1}{c|}{\textit{18.6\%}}          & 0.231          & \multicolumn{1}{c|}{\textit{25.4\%}}          & \multicolumn{1}{c|}{0.908}             & 2.095        \\
\multicolumn{1}{c|}{J2-J3}                               & \textbf{0.104} & \multicolumn{1}{c|}{\textit{\textbf{12.1\%}}} & \textbf{0.118} & \multicolumn{1}{c|}{\textit{\textbf{13.7\%}}} & \textbf{0.174} & \multicolumn{1}{c|}{\textit{\textbf{20.2\%}}} & \multicolumn{1}{c|}{0.859}             & 2.103        \\
\multicolumn{1}{c|}{J1-J3}                               & 0.120          & \multicolumn{1}{c|}{\textit{12.8\%}}          & \textbf{0.114} & \multicolumn{1}{c|}{\textit{\textbf{12.3\%}}} & 0.257          & \multicolumn{1}{c|}{\textit{27.5\%}}          & \multicolumn{1}{c|}{0.932}             & 2.125        \\
\multicolumn{1}{c|}{J1-J2-J3}                            & 0.143          & \multicolumn{1}{c|}{\textit{16.5\%}}          & 0.145          & \multicolumn{1}{c|}{\textit{16.7\%}}          & 0.221          & \multicolumn{1}{c|}{\textit{25.5\%}}          & \multicolumn{1}{c|}{0.866}             & 2.219        \\ \hline
\multicolumn{9}{c}{Joint 2 (deg)}                                                                                                                                                                                                                                                                                   \\ \hline
\multicolumn{1}{c|}{J1}                                  & 0.248          & \multicolumn{1}{c|}{\textit{22.4\%}}          & 0.311          & \multicolumn{1}{c|}{\textit{28.1\%}}          & 0.288          & \multicolumn{1}{c|}{\textit{26.1\%}}          & \multicolumn{1}{c|}{1.106}             & 7.927        \\
\multicolumn{1}{c|}{J2}                                  & 0.313          & \multicolumn{1}{c|}{\textit{27.6\%}}          & 0.242          & \multicolumn{1}{c|}{\textit{21.3\%}}          & 0.250          & \multicolumn{1}{c|}{\textit{22.0\%}}          & \multicolumn{1}{c|}{1.135}             & 8.016        \\
\multicolumn{1}{c|}{J3}                                  & 0.241          & \multicolumn{1}{c|}{\textit{20.8\%}}          & \textbf{0.203} & \multicolumn{1}{c|}{\textit{\textbf{17.5\%}}} & 0.294          & \multicolumn{1}{c|}{\textit{25.4\%}}          & \multicolumn{1}{c|}{1.160}             & 8.115        \\
\multicolumn{1}{c|}{J1-J2}                               & 0.234          & \multicolumn{1}{c|}{\textit{20.1\%}}          & 0.276          & \multicolumn{1}{c|}{\textit{23.6\%}}          & 0.221          & \multicolumn{1}{c|}{\textit{19.0\%}}          & \multicolumn{1}{c|}{1.168}             & 8.064        \\
\multicolumn{1}{c|}{J2-J3}                               & \textbf{0.120} & \multicolumn{1}{c|}{\textit{\textbf{10.9\%}}} & 0.222          & \multicolumn{1}{c|}{\textit{20.3\%}}          & \textbf{0.186} & \multicolumn{1}{c|}{\textit{\textbf{17.0\%}}} & \multicolumn{1}{c|}{1.095}             & 7.963        \\
\multicolumn{1}{c|}{J1-J3}                               & 0.265          & \multicolumn{1}{c|}{\textit{22.7\%}}          & 0.294          & \multicolumn{1}{c|}{\textit{25.2\%}}          & 0.266          & \multicolumn{1}{c|}{\textit{22.8\%}}          & \multicolumn{1}{c|}{1.168}             & 7.985        \\
\multicolumn{1}{c|}{J1-J2-J3}                            & \textbf{0.130} & \multicolumn{1}{c|}{\textit{\textbf{12.6\%}}} & \textbf{0.148} & \multicolumn{1}{c|}{\textit{\textbf{14.2\%}}} & \textbf{0.169} & \multicolumn{1}{c|}{\textit{\textbf{16.2\%}}} & \multicolumn{1}{c|}{1.038}             & 7.945        \\ \hline
\multicolumn{9}{c}{Joint 3 (mm)}                                                                                                                                                                                                                                                                                    \\ \hline
\multicolumn{1}{c|}{J1}                                  & 0.179          & \multicolumn{1}{c|}{\textit{36.8\%}}          & 0.406          & \multicolumn{1}{c|}{\textit{83.5\%}}          & 0.239          & \multicolumn{1}{c|}{\textit{49.2\%}}          & \multicolumn{1}{c|}{0.486}             & 11.723       \\
\multicolumn{1}{c|}{J2}                                  & 0.216          & \multicolumn{1}{c|}{\textit{34.4\%}}          & 0.357          & \multicolumn{1}{c|}{\textit{56.9\%}}          & 0.257          & \multicolumn{1}{c|}{\textit{40.9\%}}          & \multicolumn{1}{c|}{0.628}             & 11.737       \\
\multicolumn{1}{c|}{J3}                                  & \textbf{0.111} & \multicolumn{1}{c|}{\textit{\textbf{22.4\%}}} & \textbf{0.141} & \multicolumn{1}{c|}{\textit{\textbf{28.3\%}}} & \textbf{0.134} & \multicolumn{1}{c|}{\textit{\textbf{26.9\%}}} & \multicolumn{1}{c|}{0.498}             & 11.754       \\
\multicolumn{1}{c|}{J1-J2}                               & 0.183          & \multicolumn{1}{c|}{\textit{31.3\%}}          & 0.416          & \multicolumn{1}{c|}{\textit{71.3\%}}          & 0.253          & \multicolumn{1}{c|}{\textit{43.4\%}}          & \multicolumn{1}{c|}{0.584}             & 11.773       \\
\multicolumn{1}{c|}{J2-J3}                               & \textbf{0.118} & \multicolumn{1}{c|}{\textit{\textbf{24.0\%}}} & \textbf{0.173} & \multicolumn{1}{c|}{\textit{\textbf{35.1\%}}} & \textbf{0.166} & \multicolumn{1}{c|}{\textit{\textbf{33.7\%}}} & \multicolumn{1}{c|}{0.492}             & 11.715       \\
\multicolumn{1}{c|}{J1-J3}                               & 0.161          & \multicolumn{1}{c|}{\textit{29.0\%}}          & 0.178          & \multicolumn{1}{c|}{\textit{31.9\%}}          & 0.179          & \multicolumn{1}{c|}{\textit{32.2\%}}          & \multicolumn{1}{c|}{0.557}             & 11.778       \\
\multicolumn{1}{c|}{J1-J2-J3}                            & 0.164          & \multicolumn{1}{c|}{\textit{34.9\%}}          & 0.274          & \multicolumn{1}{c|}{\textit{58.5\%}}          & 0.168 & \multicolumn{1}{c|}{\textit{35.8\%}} & \multicolumn{1}{c|}{0.469}             & 11.703       \\ \hline
\multicolumn{9}{l}{* i) The best 2 directions of calibration trajectories for each joint and calibration model are marked bold.}\\
\multicolumn{9}{l}{\: ii) The percentage is computed based on the fixed offset compensation instead of raw errors.}\\

\end{tabular}
\end{center}
\vspace{-2.5em}
\label{tab_diff_dir}
\end{table*}
As described in \ref{sch_wkspc_n_traj} and Fig. \ref{fig_diff_trajs}, different directions of calibration trajectories cover the joints with different sparsity. Thus, the performance of the calibration was evaluated using trajectories with 7 different directions. For each direction, 3 calibration zig-zag trajectories with sparsity of $\frac{1}{2}$, $\frac{1}{3}$, $\frac{1}{4}$ were collected, with an independent 20-minute random movement trajectory as the test trajectory. All calibrations were performed using the combination of the 3 zig-zag trajectories. Evaluation of the DNN model was the average 5 repetitive training and evaluations with different global random seeds, which applied to all evaluations of DNN models in this paper.

The calibration accuracy (RMSE) of different models trained by zig-zag trajectories with different directions is shown in Table \ref{tab_diff_dir}. For joint 1, fixed offset compensation reduced the raw error from $\sim2^{\circ}$ to $\sim0.9^{\circ}$. The best performance of DNN calibration was achieved by the direction J2-J3 of the training trajectory, with an accuracy of $0.104^{\circ}$. Compared to fixed offset compensation, DNN calibration further reduced $87.9\%$ of the error. The linear regression model also had the best accuracy of $0.174^{\circ} (-79.8\%)$ trained with the same trajectory direction. 2nd-order polynomial regression model achieved its second-best accuracy of $0.118^{\circ} (-86.3\%)$, while the best accuracy $0.114^{\circ} (-87.7\%)$ was trained by direction J1-J3 for this model.

For joint 2, the fixed offset compensation reduced the raw error from $\sim8.0^{\circ}$ to $\sim1.1^{\circ}$. The best accuracy $0.120^{\circ} (-89.1\%)$ was achieved by the DNN model trained with direction J2-J3. When trained with direction J1-J2-J3, the 2-nd polynomial regression and linear regression models saw their best accuracy $0.148^{\circ} (-85.8\%)$ and $0.169^{\circ} (-83.8\%)$, respectively. 

Similarly, for joint 3, the fixed offset compensation reduced the raw error from $\sim11.7$ mm to $\sim0.5$ mm. The best performance of all DNN, 2nd-order polynomial regression, and linear regression was obtained by direction J3, with the accuracy of $0.111 (-77.6\%)$ mm, $0.141 (-71.7\%)$ mm, and $0.134 (-73.1\%)$ mm, respectively. The second-best performance of all models was found with direction J2-J3, with accuracy DNN -- $0.118 (-76.0\%)$ mm, 2nd-order polynomial regression -- $0.173 (-64.9\%)$ mm, and linear regression -- $0.166 (-66.3\%)$ mm. 

Overall, the direction of J2-J3 showed the best overall performance in all 3 joints among all 3 calibration models, reducing more than $64.9\%$ of error, and thus was chosen for all the rest of the experiments.

\subsection{Sparsity and Calibration Time} \label{schar_exp_sparsity}

\begin{figure*}
\centering
\vspace{0.3em}
\includegraphics[width=0.9\textwidth]{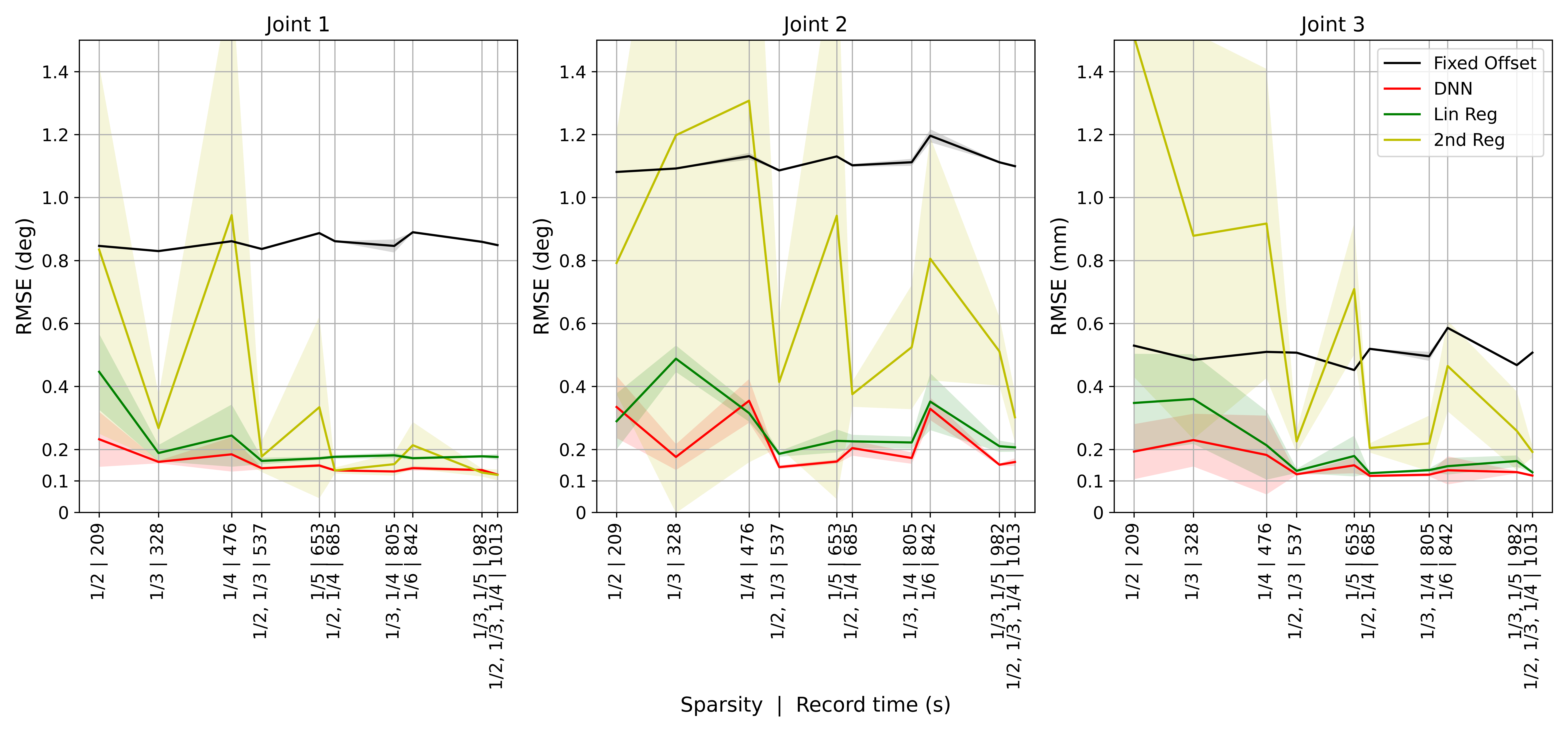}
\vspace{-1.em}
\caption{Calibration performance with different sparsities of the calibration trajectory. For sparsity $1/2$, $1/3$, $1/4$, 5 independent calibration trajectories were recorded, and for sparsity $1/5$, $1/6$, 3 independent were recorded to prevent long-time operating. Combinations of trajectories with different sparsities, no longer than 1013 seconds to record, were also evaluated. Average RMSE is shown in solid lines and the standard deviations are shown in shades.} 
\vspace{-1.5em}
\label{fig_exp2_sparsity}
\end{figure*}

As shown in Fig. \ref{fig_diff_trajs}, trajectories with smaller sparsities have better coverage of the robot workspace, but also require more time to collect data in the learning calibration. This experiment studied the calibration performance of the DNN and regression models trained by different trajectories, with sparsities $\frac{1}{2}$, $\frac{1}{3}$, $\frac{1}{4}$, $\frac{1}{5}$, and $\frac{1}{6}$. Combinations of different sparsities shorter than 1013 seconds were also evaluated. To rule out the time effect (details in \ref{schar_time_effect}), independent 20-minute random trajectories were collected for each sparsity as test sets. The result is shown in Fig. \ref{fig_exp2_sparsity}. For sparsities $\frac{1}{2}$, $\frac{1}{3}$, $\frac{1}{4}$, 5 independent calibration trajectories were recorded, and for sparsity $\frac{1}{5}$ and $\frac{1}{6}$, 3 independent calibration trajectories were recorded to prevent long-time running. 

For joint 1, DNN and linear regression models showed considerable improvement in all sparsities. When sparsity was $\frac{1}{2}$, it only took $209$ seconds to record the calibration trajectory. DNN calibration achieved RMSE of $0.232^{\circ}$. Compared to the RMSE of fixed offset compensation -- $0.846^{\circ}$, the linear regression model had RMSE of $0.446^{\circ}$, while 2nd-order polynomial regression showed no significant improvement -- $0.834^{\circ}$. As sparsity got smaller, the performance of all learning and regression methods was improved in both average RMSE and standard deviation. At the longest tested trajectory (1013 seconds to record), the combination of sparsities $\frac{1}{2}$, $\frac{1}{3}$ and $\frac{1}{4}$, all 3 models showed significant improvement in accuracy, DNN -- $0.120^{\circ}$, 2nd-order polynomial regression -- $0.119^{\circ}$, and linear regression -- $0.176^{\circ}$. Compared to the shortest trajectory with sparsity $\frac{1}{2}$, the longest combined trajectory had improvement in accuracy of $0.112^{\circ}$, $0.727^{\circ}$, and $0.270^{\circ}$ in DNN, 2nd-order polynomial, and linear regression, respectively. The performance of fixed offset compensation did not see improvement with smaller sparsity. Standard deviation also got reduced with sparsity in general, while a significantly larger standard deviation could be observed in 2nd-order polynomial regression when shorter calibration trajectories were used.


For joint 2, when trained by trajectories shorter than 500 seconds, the DNN model and linear regression model had an average accuracy of $0.288^{\circ}$ and $0.364^{\circ}$, respectively. On the other hand, when trained by trajectories longer than 800 seconds, the average calibration RMSE reduced to $0.203^{\circ}$ (DNN) and $0.248^{\circ}$ (linear regression). The performance of 2nd-order polynomial regression was very unstable with large standard deviations among most of the sparsities and only achieved comparable accuracy $0.302^{\circ}$ when trained by the longest trajectory, though still did not outperform the other 2 models.

For joint 3, similarly, when trained by trajectories shorter than 500 seconds, the DNN model and linear regression model had an average accuracy of $0.202$ mm and $0.307$ mm, respectively. As the calibration trajectory increased to 537 seconds, the combination of sparsities $\frac{1}{2}$ and $\frac{1}{3}$, the accuracy was improved to DNN -- $0.121$ mm and linear regression -- $0.132$ mm. No considerable further improvement could be found with longer calibration trajectories. Although the 2nd-order regression model had reduced RMSE with longer calibration trajectories, the accuracy was overall unstable and did not outperform the other 2 models.

\subsection{Long-time Idleness, Unloaded and Loaded Operation} \label{schar_time_effect}

\begin{figure*}
\centering
\vspace{0.3em}
\includegraphics[width=0.9\textwidth]{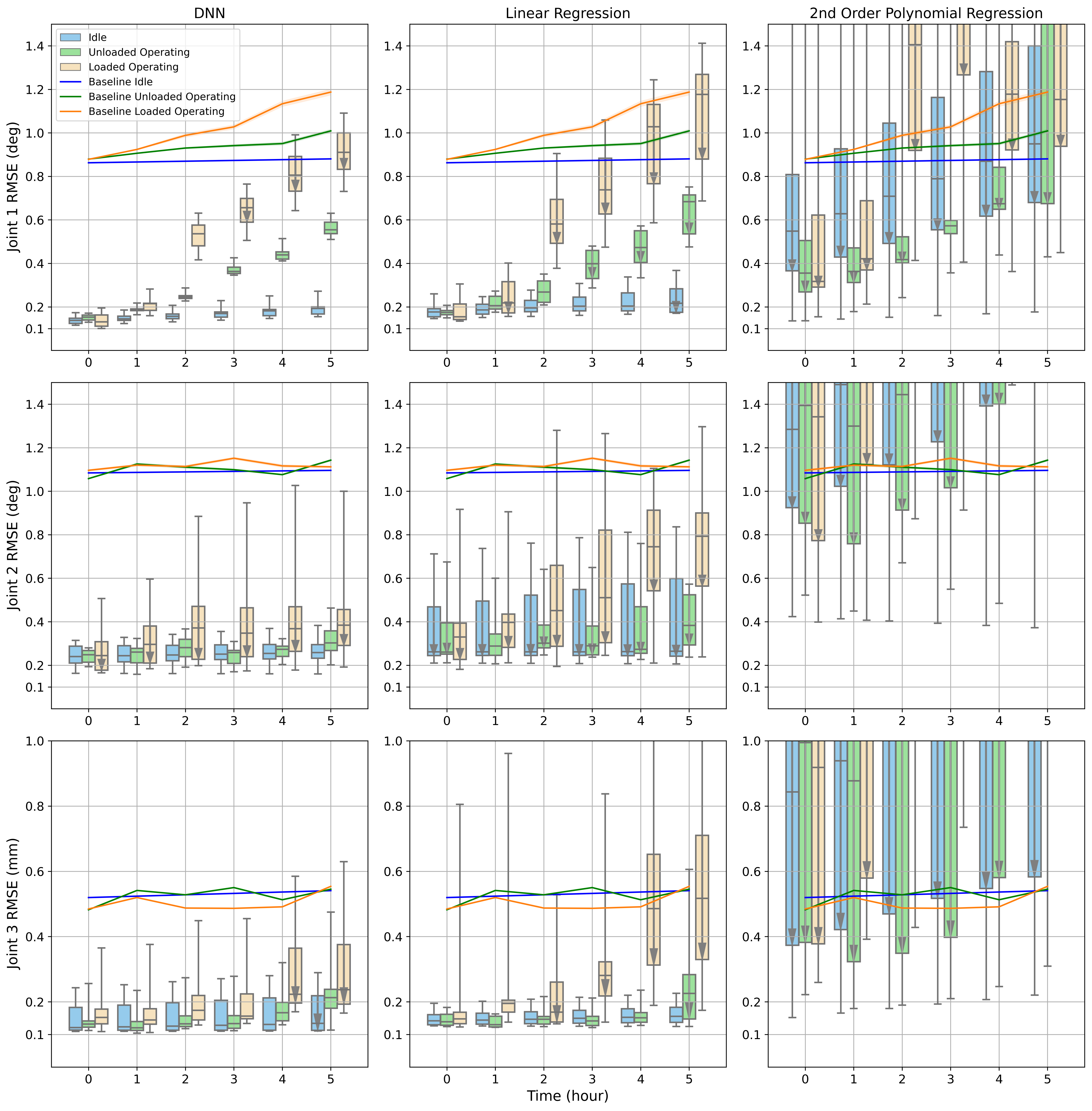}
\vspace{-1.em}
\caption{Calibration performance during 6-hour continuous unloaded operating, 500g loaded operating, and idleness. The same sparsities were used as in \ref{schar_exp_sparsity} and shown in the box plot. The baseline results were obtained by fixed offset compensation. Accuracy of idleness was only evaluated by test trajectories in hours 0 and 5, and hours 1-4 were interpolated. The arrows in the box plot indicate training trajectories longer than 800 sec had $>$20\% better calibration accuracy than trajectories shorter than 500 sec, and the difference between the 1st and 3rd quartile is larger than 0.2 (deg/mm).} 
\vspace{-1.5em}
\label{fig_exp3_time_decay}
\end{figure*}

\begin{figure*}
\centering
\vspace{0.3em}
\includegraphics[width=0.9\textwidth]{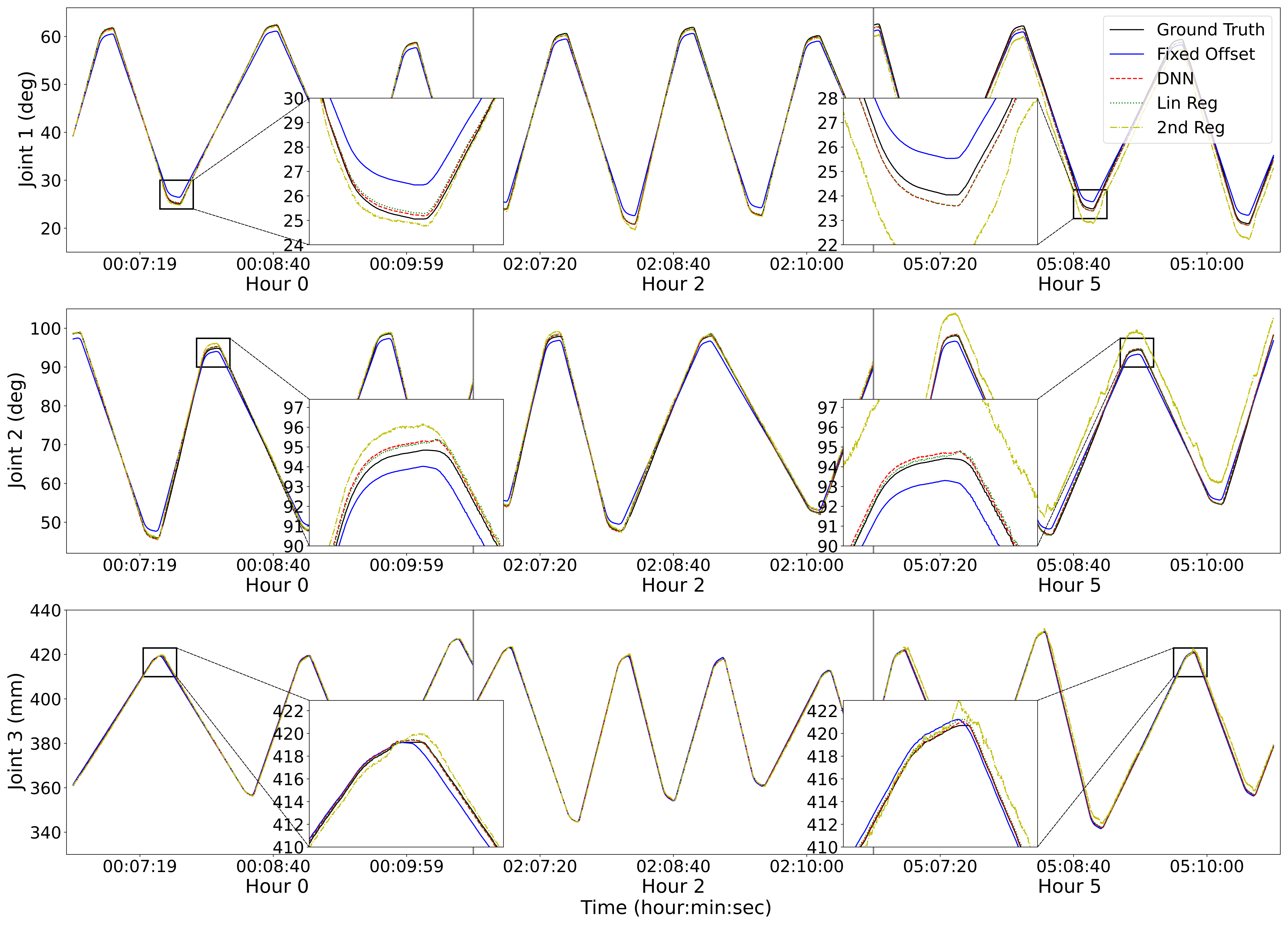}
\vspace{-1.em}
\caption{Example of calibrated joint position trajectories. The models were trained by the calibration trajectory with sparsity $\frac{1}{6}$. 3 sections are shown horizontally. The first section is the immediate performance right after calibration. The second and third sections show performance in hour 2 and 5, respectively.} 
\vspace{-1.5em}
\label{fig_exp3_cali_rst}
\end{figure*}

The previous experiments \ref{schar_exp_dir} and \ref{schar_exp_sparsity} showed that data-driven calibrations significantly reduced the error of the joint positions. However, the feasibility of the calibrations also depends on how the calibration accuracy drops with operating time. Due to the relatively small sheave and cable diameter ratio, high tension, large wrap angle, etc., cables of RAVEN-II have a short fatigue life and cable characteristics can change over time \cite{miyasaka2017cable}. To study the influence of operating time, calibration trajectories in the direction of J2-J3 with different sparsities were recorded first. Then the robot was continuously operated by random trajectories for 6 hours to collect the test set. Unloaded and loaded (500g weight scale hanging on the end-effector) configurations were tested, as well as unloaded idleness for comparison. The result is shown in Fig. \ref{fig_exp3_time_decay}. The same sparsities were used as in \ref{schar_exp_sparsity} and shown in the box plot. The arrows in the box plot indicate training trajectories longer than 800 sec had $>$20\% better calibration accuracy than trajectories shorter than 500 sec, which suggests considerably better performance can be obtained by longer calibration trajectories. As shown in Fig. \ref{fig_exp3_time_decay}, longer calibration trajectories achieved better performance in many entries, especially for regression models. The RMSE described in this section is the average RMSE of models trained by the 3 longest trajectories, unless stated otherwise.

For joint 1, for idleness of 6 hours, the baseline result -- fixed offset compensation had no obvious decrease in accuracy from hour 0 -- $0.863^{\circ}$ to hour 5 -- $0.879^{\circ}$. The DNN model saw a small drop of $0.048^{\circ}$ in accuracy from hour 0 -- $0.124^{\circ}$ to hour 5 -- $0.172^{\circ}$. A similar trend was observed in the linear regression model, from $0.162^{\circ}$ to $0.193^{\circ}$. However, the 2nd-order polynomial regression model saw a significant decrease in accuracy, from $0.241^{\circ}$ to $0.493^{\circ}$ even when the robot stayed idle for 6 hours. When the robot was operated for 6 hours without load, a considerable drop in accuracy could be found in all 3 calibration models. Compared to fixed offset compensation that dropped $0.121^{\circ}$ in 6 hours (from $0.879^{\circ}$ to $1.000^{\circ}$), the accuracy of the DNN model dropped $0.411^{\circ}$ (from $0.142^{\circ}$ to $0.553^{\circ}$), the linear regression model dropped $0.358^{\circ}$ (from $0.169^{\circ}$ to $0.527^{\circ}$), and the 2nd-order polynomial regression model dropped $1.294^{\circ}$ (from $0.235^{\circ}$ to $1.529^{\circ}$). It is also worth noticing that after hour 3, longer trajectories could result in better performance in the linear regression model, while the DNN model is less affected by the amount of training data. When the robot was operated for 6 hours with a 500g load on the end-effector, a larger decay in accuracy could be found in all models. The fixed offset compensation had accuracy dropped $0.296^{\circ}$ (from $0.878^{\circ}$ to $1.174^{\circ}$), the DNN model dropped $0.741^{\circ}$ (from $0.114^{\circ}$ to $0.855^{\circ}$), the linear regression model dropped $0.772^{\circ}$ (from $0.144^{\circ}$ to $0.916^{\circ}$), and the 2nd-order polynomial regression model dropped $0.824^{\circ}$ (from $0.221^{\circ}$ to $1.045^{\circ}$).

For joint 2, with 6-hour idleness, no decay in accuracy could be found in fixed offset compensation -- from $1.084^{\circ}$ to $1.094^{\circ}$, DNN model -- from $0.211^{\circ}$ to $0.217^{\circ}$, and linear regression model -- from $0.237^{\circ}$ to $0.232^{\circ}$. The performance 2nd-order polynomial regression model was much worse and unstable than the other calibration models in all configurations and thus will not be compared. When operated for 6 hours without load, the drops in accuracy were: fixed offset compensation -- $0.083^{\circ}$ (from $1.056^{\circ}$ to $1.139^{\circ}$), DNN -- $0.099^{\circ}$ (from $0.232^{\circ}$ to $0.331^{\circ}$), linear regression -- $0.106^{\circ}$ (from $0.247^{\circ}$ to $0.353^{\circ}$). When operated for 6 hours with 500g load, the drop in accuracy was: fixed offset compensation -- $0.017^{\circ}$ (from $1.092^{\circ}$ to $1.109^{\circ}$), DNN -- $0.181^{\circ}$ (from $0.251^{\circ}$ to $0.432^{\circ}$), linear regression -- $0.322^{\circ}$ (from $0.266^{\circ}$ to $0.588^{\circ}$). It could be found that although the fixed offset compensation did not see a big difference, a larger drop in accuracy was seen by the DNN and linear regression models. Overall, in the configurations of idleness and unloaded operation, the DNN model achieved desirable performance in all calibration trajectories, while longer trajectories were preferred by the linear regression model to achieve comparable accuracy. In the loaded configuration, longer calibration trajectories were preferred by both models.

For joint 3, with 6-hour idleness, no decay in accuracy could be found in fixed offset compensation -- from $0.520$ mm to $0.541$ mm, DNN model -- from $0.113$ mm to $0.113$ mm, and linear regression model -- from $0.131$ mm to $0.132$ mm. The performance 2nd-order polynomial regression model was again excluded for comparison because of undesirable performance. When operated for 6 hours without load, the drop in accuracy was: fixed offset compensation -- $0.063$ mm (from $0.482$ mm to $0.545$ mm), DNN -- $0.039$ mm (from $0.126$ mm to $0.165$ mm), linear regression -- $0.068$ mm (from $0.132$ mm to $0.200$ mm). When operated for 6 hours with a 500g load, the drop in accuracy was: fixed offset compensation -- $0.069$ mm (from $0.485$ mm to $0.554$ mm), DNN -- $0.059$ mm (from $0.122$ mm to $0.181$ mm), linear regression -- $0.335$ mm (from $0.138$ mm to $0.473$ mm). Both DNN and linear regression models showed desirable accuracy with all calibration trajectories in the configuration of idleness and unloaded operation. However, for loaded operation, longer calibration trajectories were necessary to obtain better accuracy. 

Overall, with sufficient training data, the calibration models improved the accuracy significantly among 6-hour idleness, unloaded, and loaded operating. Examples of calibrated joint position trajectories in the test set can be found in Fig. \ref{fig_exp3_cali_rst}.

\subsection{Robustness of Unnecessary Input Features}\label{schar_robust_features}

As described in \ref{sch_dnn_reg_models}, there are numerous types of information in the robot state (details in \ref{schar_robot_states}), including the robot run level, joint position, velocity, motor torque, end-effector pose, and so on. Among these types of information, based on the ablation study \cite{peng2023ablation}, only joint positions and motor torques were found important to the calibration. However, the ablation study also took a considerable effort. Without the ablation study, in order to not miss critical input features, all information in the robot state should be given as input to the calibration models, which will increase the parameters of the models. Under the same amount of training data, more parameters may result in a higher risk of overfitting. Thus, this experiment utilized the same 6-hour unloaded operation data in \ref{schar_time_effect}. The DNN, linear regression, and 2nd-order regression models were trained using only selected useful features (joint positions and motor torques, dimension of 16), and using all features in the robot state (dimension of 138). Besides the original DNN model (2 layers, 100 hidden units each), a larger DNN model was also used to accommodate more input dimensions, which has 3 layers with 600, 500, and 400 units, respectively. Regularization rates were also applied, kernel L1 $10^{-5}$ L2 $10^{-4}$, bias L2 $10^{-4}$, and activity L2 $10^{-5}$. The other hyperparameters stayed unchanged. GPU acceleration was utilized for the training of the larger DNN model. 

\begin{figure*}
\centering
\vspace{0.3em}
\includegraphics[width=0.8\textwidth]{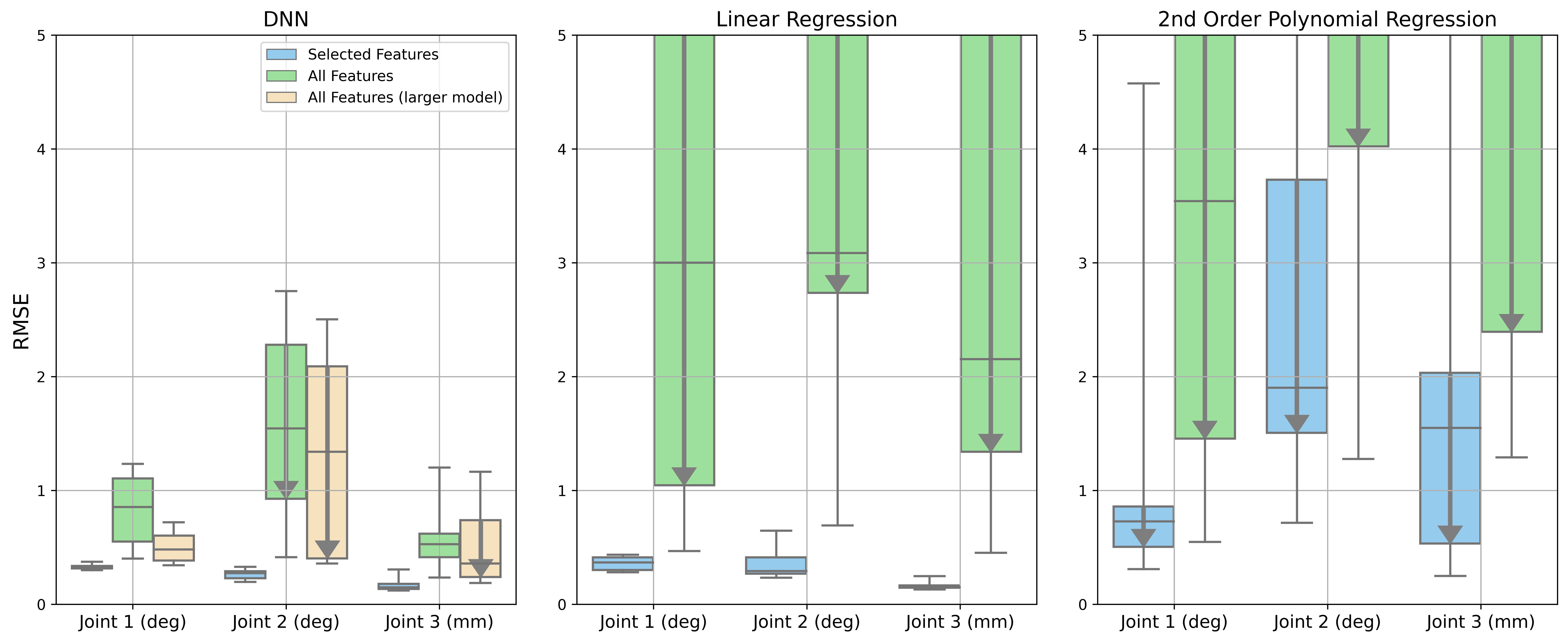}
\vspace{-1.em}
\caption{Calibration performance of models trained by selected important features (joint positions and motor torques), and by all features (details in \ref{schar_robot_states}) in the robot state. Besides the original DNN model, a larger DNN model was also used to accommodate more input dimensions. The boxed plot shows the average RMSE among 6-hour unloaded operating with different calibration trajectory sparsities, as in Fig. \ref{fig_exp3_time_decay}. The arrows in the box plot indicate training trajectories longer than 800 sec had $>$20\% better calibration accuracy than trajectories shorter than 500 sec, and the difference between the 1st and 2nd quartile is larger than 0.2 (deg/mm).} 
\vspace{-1.5em}
\label{fig_exp5_feature_selection}
\end{figure*}

The evaluation result is shown in Fig. \ref{fig_exp5_feature_selection}. When the model input was only joint positions and motor torques, the DNN model had desirable average accuracy in 6-hour unloaded operating among all calibration trajectories from short to long, with the RMSE of joint 1 -- $0.328^{\circ}$, joint 2 -- $0.264^{\circ}$ and joint 3 -- $0.167$ mm. When all features in the robot state were used as input, the input dimension was 8.6 times larger from 16 to 138, but the number of model weight parameters was only 2 times larger from 12103 to 24303. A considerable drop in accuracy could be observed in all 3 joints. The average RMSE when trained by the 3 longest calibration trajectories was joint 1 -- $0.700^{\circ}$, joint 2 -- $0.754^{\circ}$, and joint 3 -- $0.388$ mm. When the 3-layer larger model was used with all features as input, the weight parameters increased dramatically to 585503. The average RMSE when trained by 3 longest calibration trajectories improved to joint 1 -- $0.397^{\circ}$, joint 2 -- $0.565^{\circ}$, and joint 3 -- $0.229$ mm, though the performance was still not as good as the small model with selected features.

The linear regression model had 17 parameters for each joint when trained by selected features, and had desirable average accuracy in 6-hour unloaded operating among all calibration trajectories, with the RMSE of joint 1 -- $0.361^{\circ}$, joint 2 -- $0.355^{\circ}$ and joint 3 -- $0.167$ mm. When all 138 features in the robot state were used as input, the number of parameters also increased linearly to 139. A significant drop in accuracy could be observed, and the average RMSE trained by the 3 longest calibration trajectories was joint 1 -- $1.985^{\circ}$, joint 2 -- $1.834^{\circ}$, and joint 3 -- $1.352$ mm, which did not outperform fixed offset compensation. A similar trend could also be observed by the 2nd-order regression model, but the performance was worse than linear regression. 

\subsection{Effects of Homing}

\begin{figure*}
\centering
\vspace{0.3em}
\includegraphics[width=0.8\textwidth]{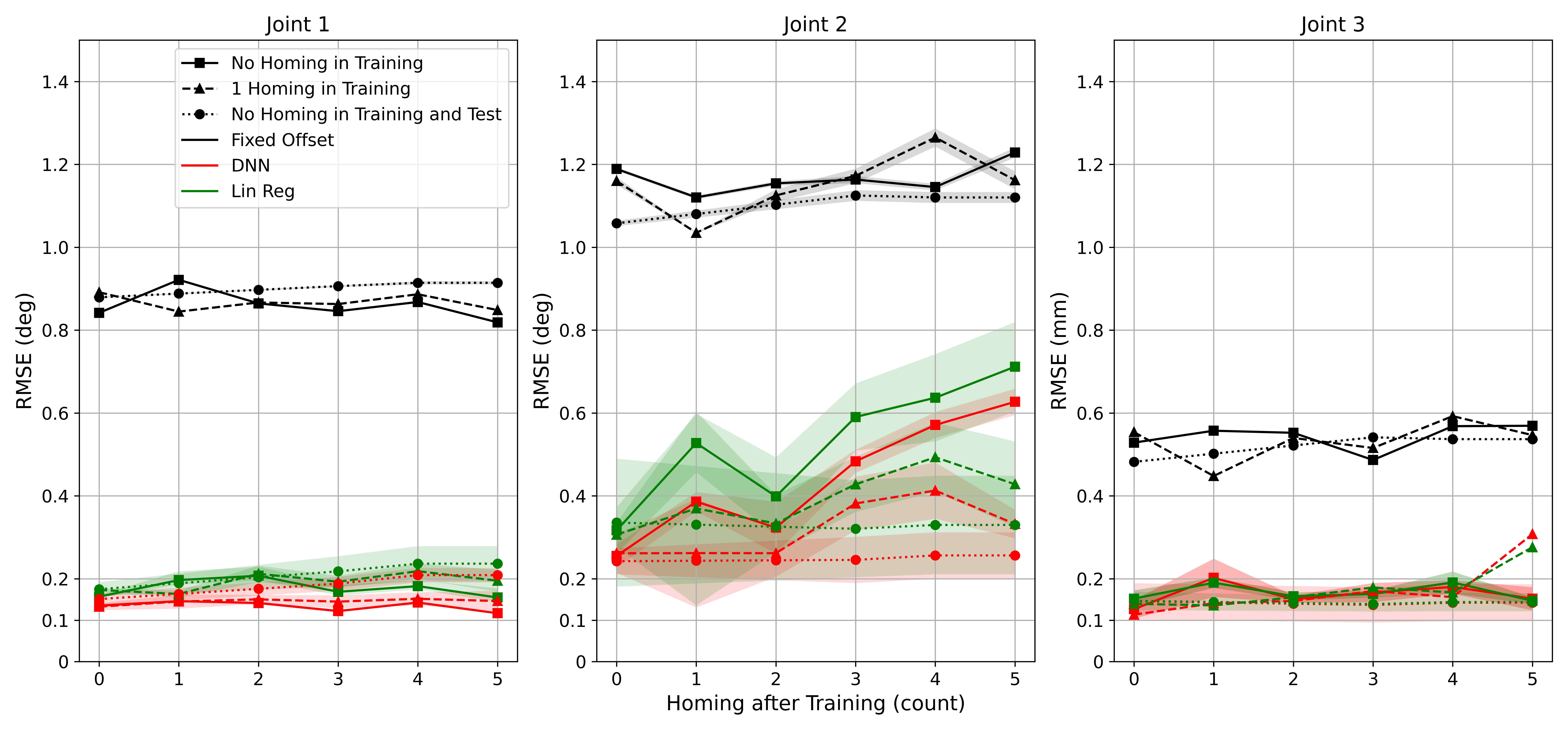}
\vspace{-1.em}
\caption{Calibration performance after 0-5 robot homing procedures. Different combinations (\textless20 min) of training trajectories with various sparsities were tested and the average is shown with standard deviation as shade. The 2nd order polynomial regression model is not shown because of unstable performance. Two datasets were used for training, one had the training set with no homing, and another one had the training set with 1 homing. To study the effectiveness of homing, baseline results with the same operating time but no homing in both training and test sets are also included.} 
\vspace{-1.5em}
\label{fig_exp4_homing_decay}
\end{figure*}

The homing procedure is an important function of the RAVEN-II robot, during which the robot explores the hard limits of all the joints and registers the encoders on the motors. Joint positions will be meaningful after the homing procedure. Re-homing may sometimes be necessary during operations, for example, the surgical instrument is changed or software e-stop is triggered because of joint limits or singularity. Thus, it is important to study whether the homing procedure influences the calibration's performance. Two new datasets were collected to study the effectiveness of homing. In each dataset, 5 homings were performed and a 20-minute random trajectory was recorded as test data before the first homing, in between the homings, and after the last homing. The difference between the two datasets was that one dataset had all training calibration trajectories before the first homing, thus this training set 'saw' no homing. The other dataset had the same sparsities and length of the training trajectories, but these trajectories were separated before and after the first homing, thus this training set 'saw' 1 homing.

The evaluation of calibration performance among homings is shown in Fig. \ref{fig_exp4_homing_decay}. For joint 1, the accuracy of the DNN model slightly fluctuated between $0.117^{\circ}$ and $0.146^{\circ}$ among 5 homings with no homing 'seen' in training. With 1 homing 'seen' in training, the accuracy of the DNN model slightly fluctuated between $0.133^{\circ}$ and $0.152^{\circ}$. No obvious drop in accuracy could be observed caused by homing. A similar trend was found using the linear regression model, fluctuating between $0.155^{\circ}$ and $0.207^{\circ}$ with no homing in training, and fluctuating between $0.163^{\circ}$ and $0.218^{\circ}$ with 1 homing in training.

For joint 2, when the calibration models 'saw' no homing in training, the accuracy of the DNN model dropped significantly from $0.255^{\circ}$ to $0.627^{\circ}$ after 5 homings, and similarly, the accuracy of the linear regression dropped from $0.317^{\circ}$ to $0.712^{\circ}$. On the other hand, if the models 'saw' 1 homing in training, although the accuracy still dropped with homing, the RMSE stayed within $0.413^{\circ}$ for the DNN model and within $0.493^{\circ}$ among 5 homings. 

For joint 3, except for an outlier at the 5th homing, the DNN model and the linear regression model had close accuracy regardless of 'seeing' homing in training or not, slightly fluctuating from $0.113$ mm to $0.202$ mm.

\subsection{Training on Error}\label{schar_train_on_err}
As introduced in \ref{sch_dnn_reg_models} and Fig. \ref{fig_learn_reg_model}, the DNN and regression models were not trained as end-to-end models that directly outputted the calibrated joint positions. Instead, the models outputted the errors of the inaccurate original joint positions in the robot state. The errors were then used to correct the original joint positions and thus provided calibrated accurate joint positions. To study the necessity of training on error, this experiment used the same 6-hour unloaded data as \ref{schar_time_effect}, and the models were trained on error or end-to-end. 

The comparison of calibration accuracy is shown in Table. \ref{tab_train_on_err}. For the regression models, no difference in RMSE could be found between models trained on error or end-to-end, though the average performance of 2nd-order polynomial regression was not desirable. On the other hand, for the DNN model, when trained on error, the best accuracy was achieved, joint 1 -- $0.328^{\circ}$, joint 2 -- $0.264^{\circ}$, and joint 3 -- $0.167$ mm. However, when trained end-to-end, the calibration error increased sharply to $0.830^{\circ}$, $0.875^{\circ}$, and $1.583$ mm in joints 1, 2, and 3, respectively. Further tests found that increasing the number of training epochs could improve the performance of the end-to-end DNN model, however, the accuracy was still not as good as the DNN model trained on error. Since training on error requires no extra effort and can be faster with fewer epochs, it is highly preferred for efficient calibration.

\renewcommand{\arraystretch}{1.5}
\begin{table*}[h]
\caption{\vspace{-0.3em} Calibration Performance of Models Trained on Joint Position Error or End-to-end}
\vspace{-1.5em}
\begin{center}
\begin{tabular}{c|cc|cc|cc}
\hline
\multirow{2}{*}{Model} & \multicolumn{2}{c|}{Joint 1 (deg)} & \multicolumn{2}{c|}{Joint 2 (deg)} & \multicolumn{2}{c}{Joint 3 (mm)} \\
                       & Err              & E-to-E          & Err              & E-to-E          & Err             & E-to-E         \\ \hline
DNN                    & \textbf{0.328}   & \textbf{0.831}  & \textbf{0.264}   & \textbf{0.875}  & \textbf{0.167}  & \textbf{1.583} \\
Lin Reg                & 0.369            & 0.369           & 0.366            & 0.366           & 0.176           & 0.176          \\
Poly Reg               & 1.062            & 1.062           & 2.539            & 2.539           & 1.766           & 1.766          \\ \hline
\multicolumn{7}{l}{* Each entry is the average RMSE of all sparsities among 6 hours.}\\
\end{tabular}
\end{center}
\label{tab_train_on_err}
\end{table*}

\subsection{Comparison to the State-of-the-art}
The comparison of calibration performance with the state-of-the-art in the joint space is shown in Table \ref{tab_direct_SOTA}, and the comparison in Cartesian space is shown in Table. \ref{tab_indirect_SOTA}. The proposed method in this paper performs calibration in the first 3 joints. Thus, the location accuracy in Cartesian space is obtained by assuming the last 4 joints are accurate and the surgical tool has no deformation. The last 4 joints of RAVEN-II are concentrated near the end-effector with only one non-zero link length of 13 mm. Thus, the inaccuracy of the last 4 joints mainly results in errors in the orientation of the end-effector but less in position. However, a certain error increment still exists in real practice because of deformation.

Compared with the state-of-the-art works, we also focus on efficiency while achieving state-of-the-art accuracy on RAVEN-II. We trained the calibration models within 21 minutes and evaluated the performance of calibration in 6-hour unloaded and loaded operating, and among robot re-homings, in order to prove the feasibility of applying data-driven calibration in the time scale of real surgeries. Besides using computationally expensive DNN models that have slower inference time, by excluding unnecessary input features \cite{peng2023ablation} and providing sufficient training data, we also show that the linear regression model can have a significantly faster inference speed that can be applied to the 1000 Hz servo control of RAVEN-II, with slightly compromised accuracy.

\begin{table*}[h]
\caption{Direct Comparison to the State-of-the-art in Joint Space}
\begin{center}
\begin{tabular}{c|c|c|c|c|c}
\hline
Method & Robot & Model & Joint 1 (deg) & Joint 2 (deg) & Joint 3 (mm) \\ \hline
Ours (hour 0--5) & RAVEN-II & DNN & 0.142--0.553 & 0.230--0.343 & 0.125--0.176 \\
Ours (hour 0--5) & RAVEN-II & Lin Reg & 0.169--0.522 & 0.248--0.376 & 0.133--0.206 \\
Haghighipanah et al. \cite{haghighipanah2015improving} & RAVEN-II & UKF & 1.195 & 1.242 & 0.927 \\ 
Hwang et al. \cite{hwang2022automating} (left arm) & dVRK & RNN & 0.115 & 0.069 & 0.150 \\ 
Hwang et al. \cite{hwang2022automating} (right arm) & dVRK & RNN & 0.172 & 0.206 & 0.510 \\ 
\hline
\multicolumn{6}{l}{* Performances of unloaded configuration is presented in this table.}\\

\end{tabular}
\end{center}
\label{tab_direct_SOTA}
\end{table*}

\begin{table*}[h]
\caption{Indirect Comparison to the State-of-the-art in Cartesian Space}
\begin{center}
\begin{tabular}{c|c|c|c|c|c|c}
\hline
Method & Robot & Model & X (mm) & Y (mm) & Z (mm) & Dist (mm) \\ \hline
Ours* (hour 0--5, est.) & RAVEN-II & DNN & 0.362--0.538 & 0.301--1.503 & 0.205--0.581 & 0.513--1.318 \\
Ours* (hour 0--5, est.) & RAVEN-II & Lin Reg & 0.389--0.589 & 0.351--0.998 & 0.231--0.564 & 0.572--1.289 \\
Hwang et al. \cite{hwang2022automating} (pick) & dVRK & RNN & -- & -- & -- & 0.65 \\
Hwang et al. \cite{hwang2022automating} (random) & dVRK & RNN & -- & -- & -- & 2.05 \\
Mahler et al. \cite{mahler2014learning} & RAVEN-II & GPR & 1.4 & 1.4 & 1.1 & -- \\
Peng et al. \cite{peng2020real} & RAVEN-II & DNN & 0.818 & 1.078 & 0.518 & -- \\ \hline

\multicolumn{7}{l}{* The performance of our proposed method was estimated by assuming the last 4 joints were accurate and there}\\
\multicolumn{7}{l}{\:\:\: was no deformation of the surgical tool.}\\

\end{tabular}
\end{center}
\label{tab_indirect_SOTA}
\end{table*}

\section{Discussion}

\begin{table*}[h]
\caption{Deep Neural Network VS. Linear Regression}
\begin{center}
\begin{tabular}{c|l|l}
\hline
\multicolumn{1}{l|}{} & \multicolumn{1}{c|}{DNN}                            & \multicolumn{1}{c}{Linear Regression}         \\ \hline
\hline
Accuracy              & \textbf{Slightly better with sufficient data}       & Slightly worse with sufficient data           \\ \hline
Data Hunger           & \textbf{Better with small training set}             & Worse with small training set                 \\ \hline
Training Time         & \textless 50 sec, no GPU                                    & \textbf{\textless 2 sec}                       \\ \hline
Inference Time        & 61 ms, enough for CRTK interpolate level            & \textbf{0.38 ms, enough for CRTK servo level} \\ \hline
Irrelevant features      & \textbf{Robust with larger model and training set} & Not robust                                    \\ \hline
Train on Error        & Required                                            & Optional                                      \\ \hline
\end{tabular}
\end{center}
\label{tab_dnn_vs_lin}
\end{table*}

The DNN model and linear regression model showed comparable performance in the experiments reported in this paper, which raises an interesting question: is the linear regression model competitive to the DNN model? The comparison between these two models in different aspects is shown in Table. \ref{tab_dnn_vs_lin}.

In terms of accuracy, with sufficient training data, i.e. long enough calibration trajectory, the DNN model slightly outperformed the linear regression model. According to \ref{schar_time_effect}, when trained by the 3 longest calibration trajectories, the performance of the 2 models was very close in unloaded operation. When loaded, the DNN model slightly outperformed the linear regression model by $0.015^{\circ}$, $0.109^{\circ}$, and $0.124$ mm in joints 1, 2, and 3, respectively. In terms of data hunger, when trained with the 3 shortest calibration trajectories, counter-intuitively, the DNN model has considerably better performance than the linear regression model. When unloaded, the DNN model had $0.052^{\circ}$, $0.189^{\circ}$, and $-0.037$ mm better accuracy in joints 1, 2, and 3, respectively. When loaded, the difference increased to $0.128^{\circ}$, $0.360^{\circ}$, and $0.178$ mm.

With the small size of the DNN model, even with the largest training set presented in this paper, the DNN model could be trained within 50 seconds without GPU acceleration. The training time of the linear regression model was less than 2 seconds. Although the linear regression model could be trained much more quickly, compared to the entire calibration procedure described in \ref{schar_Workflow}, the data collection, which takes 4-17 minutes, remains the dominating time consumption. In terms of real-time efficiency, the inference time of the DNN model without GPU was 61 ms (\textless 20 Hz) on the Intel(R) Xeon(R) 2.2GHz CPU, which had a huge gap to meet the 1000 Hz servo control rate of the CRTK and RAVEN-II robot. However, since calibrated joint positions may not be necessary in the servo control loop of the robot, the inference time of the DNN model could meet the rate required by the CRTK's interpolate control level, which is designed for vision-based control, and so on. With GPU acceleration, the inference time could be shortened to 54 ms. The average inference time could also be significantly shortened if inferred in a mini-batch, however, this caused delay and could not be applied to real-time servo control. On the other hand, the inference time of the linear regression model was 0.38 ms (around 3500 Hz), which was sufficient for the 1000 Hz servo control rate of the robot.

As described in Section \ref{schar_robust_features} and  Fig. \ref{fig_exp5_feature_selection}, in the calibration of the RAVEN-II robot in this paper, useful features were found by an ablation study \cite{peng2023ablation}, which required a certain effort. In other cases, if useful features are unknown, in order to not miss useful features, unnecessary features may also be selected as input of the model. Under this condition, a larger DNN model showed much better robustness with a slightly compromised performance, while the linear regression model had a significant drop in accuracy and became undesirable. In terms of the model output, regression models could be either trained end-to-end (directly output calibrated joint positions) or trained on error (output the error of the original inaccurate joint positions and made correction). In comparison, the DNN model had to be trained on error, to achieve better accuracy and faster convergence. However, since no considerable extra effort was needed to perform training on error, both in the calibration phase and the real-time inference phase, there was no difference in efficiency.

To sum up, both models have their advantages and disadvantages. Depending on the use case, one may be more desirable than the other. To achieve the best accuracy, or when data collection must be fast, or unnecessary features may be given as input, the DNN calibration has better accuracy and robustness in performance. However, when calibrated joint positions are required in high frequency by real-time applications, the linear regression calibration becomes the only choice. In this case, longer calibration trajectories are critical to maintain better accuracy.


The first 3 positioning joints of the RAVEN-II robot, though all being cable-driven, have considerable differences in mechanisms (Fig. \ref{fig_raven_cable}) and calibration performance described from \ref{schar_exp_dir} to \ref{schar_train_on_err}.

Joint 1 is a single alpha-wrapped pulley joint, which has the shortest cable length and little or no free cable length. As the first joint, it also bears the largest load caused by the mass of the following robot arm links. As for calibration performance, joint 1 had the largest drop of calibration accuracy with time and load. And it was the only joint that had increased error with time and load in fixed offset compensation.

The joint 2 is a general rotational joint. Although the instantaneous accuracy right after calibration was not as good as that of joint 1, the accuracy of joint 2 dropped less with operating time and load. However, joint 2 saw a considerable increase in error with the robot homing procedure, while the other 2 joints did not.

The joint 3 is the only translational joint of the robot. To insert and extract the surgical instrument, joint 3 has a large range. However, during normal operating, only the lower half of the range is used. Approaching the higher half causes disabled inverse kinematics due to singularity. According to the feature ablation study \cite{peng2023ablation}, joint 3 was the only joint in which joint torques were more important than joint positions as input features of the calibration model. Although the DNN and linear regression models had very desirable calibration accuracy, the percentage improvement compared to fixed offset compensation was the lowest among all 3 positioning joints.

\section{Materials and Methods} \label{char_method}
\subsection{General-purpose Controller for RAVEN-II Surgical Robot} \label{sch_crtk_controller}

\begin{figure*}
\centering
\vspace{0.5em}
\includegraphics[width=0.7\textwidth]{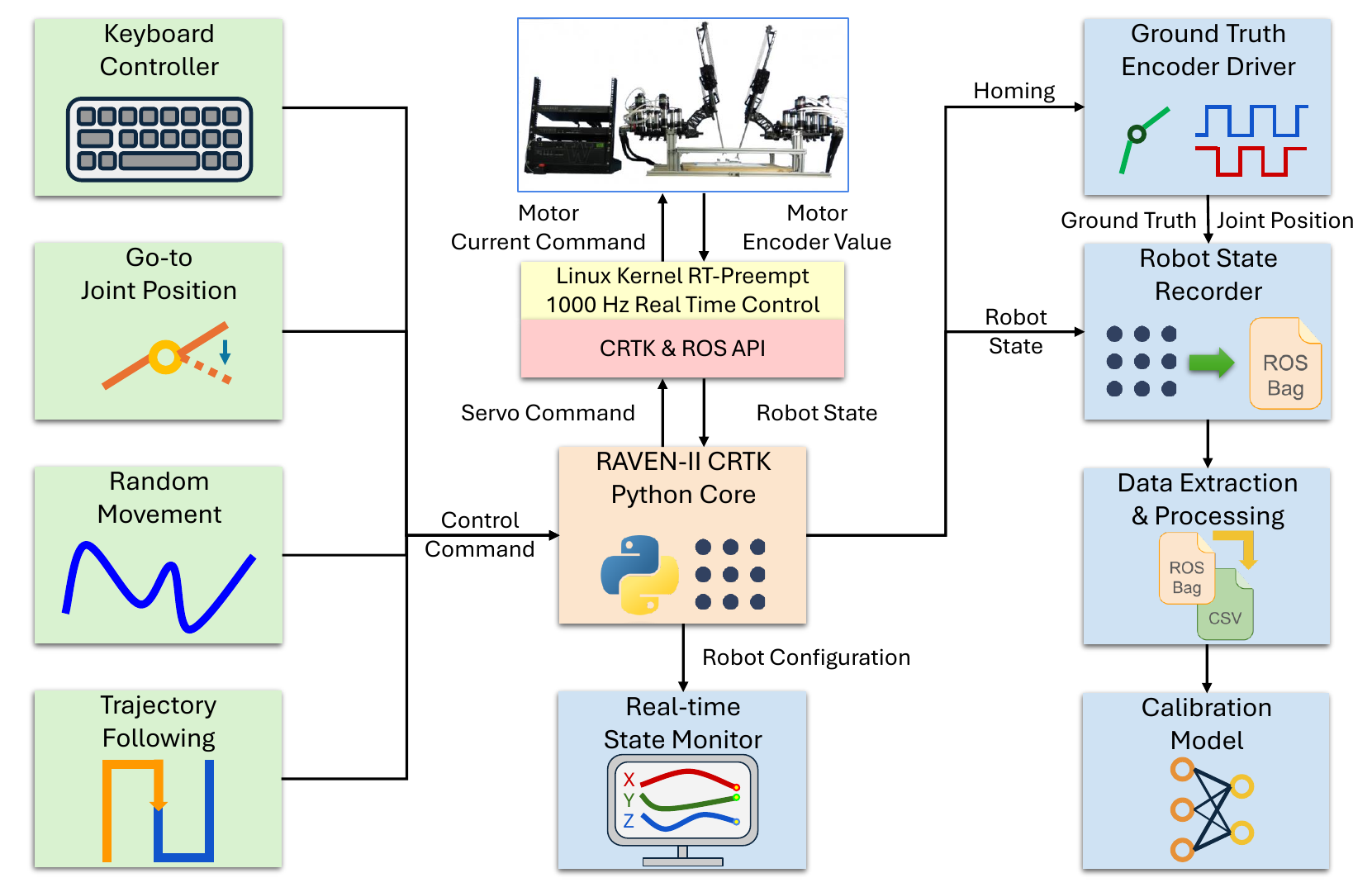}
\vspace{-1.em}
\caption{The general-purpose controller and relative tools developed for the RAVEN-II surgical robot, based on ROS, Python, and CRTK API. Compared to the original C++ controller that only supports servo incremental control, this new controller has more control, visualizations, data recording, and safety functions.} 
\vspace{-1.5em}
\label{fig_raven_controller}
\end{figure*}

To achieve efficient control and data collection on the RAVEN-II robot, a general-purpose software controller is developed, based on ROS, Python, and Collaborative Robotics Toolkit (CRTK) \cite{su2020collaborative}. CRTK is a common control API for two surgical robot research platforms, the RAVEN-II and the da Vinci Research Kit. The controller has the following parts (shown in Fig. \ref{fig_raven_controller}).

\textbf{CRTK Python core} establishes communication with the CRTK API and 'ravenstate' -- a ROS topic containing the states and measurements of the robot, including joint positions, end-effector poses, run-level, etc. It also generates and sends control commands to the robot via CRTK API. The CRTK Python core achieves semi-servo control of RAVEN-II, around 500 Hz. The control command is scaled by a factor computed in each control loop, compared to RAVEN-II's 1000Hz servo interface.

\textbf{Keyboard controller} allows the user to control RAVEN-II using local keyboard input, which enables easy testing of the functions of the robot. The robot will follow a predefined constant joint or Cartesian velocity while the corresponding key is holding pressed. The keyboard controller can also change the run level of the robot, replacing the physical pause control pedal.

\textbf{Go-to joint position} controls the robot to reach a target joint configuration, regardless of the current position. While far away from the target configuration, joints have constant velocities. The velocities get reduced while approaching the desired position, and then the joints stop within a predefined tolerance. 

\textbf{Random sinusoidal joint movement} controls the robot joints to run random sinusoidal trajectories. All joints are controlled independently. In each iteration, a target joint position is randomly chosen with a random velocity, both in predefined ranges. 


\textbf{Trajectory generation \& follow.} The purpose of this part is to control the robot to follow a target trajectory with a desired speed to collect training data for calibration. 
A trajectory is given by a sequence of configuration vectors of joint positions. The controller first moves the robot to the initial configuration using the 'go-to' control. Next, in each control loop, the current position of the robot is obtained via 'ravenstate' and CRTK API. Then, the target position is found in the trajectory, and the control command is generated based on the difference between the current position and the target position, with a desired smooth velocity.

\textbf{Runtime monitor} provides real-time visualization of the joint positions, end-effector poses, and control commands. It obtains robot states and plots them in an animated manner.

\textbf{External joint encoder}. The controller also includes an external joint encoder driver for ground truth data collection on RAVEN-II. In RAVEN-II's initialization and homing, all joints go to the limit to register motor encoders. The external joint encoders are registered in the same way and readings are then published via ROS message.

\textbf{Robot state recorder} records robot states as ROS bags via CRTK API and 'ravenstate' ROS topic, as well as external joint encoder positions. Selected CRTK joint positions and external joint positions are recorded at 100 Hz, and 'ravenstate' is recorded at 30 Hz due to high dimensions. \textbf{Data extraction} processes the ROS bags and synchronizes all robot states.

\subsection{Experiment Setup and Model Paramaters} \label{schar_exp_setup_hyper_param}

\begin{figure}
\centering
\includegraphics[width=0.45\textwidth]{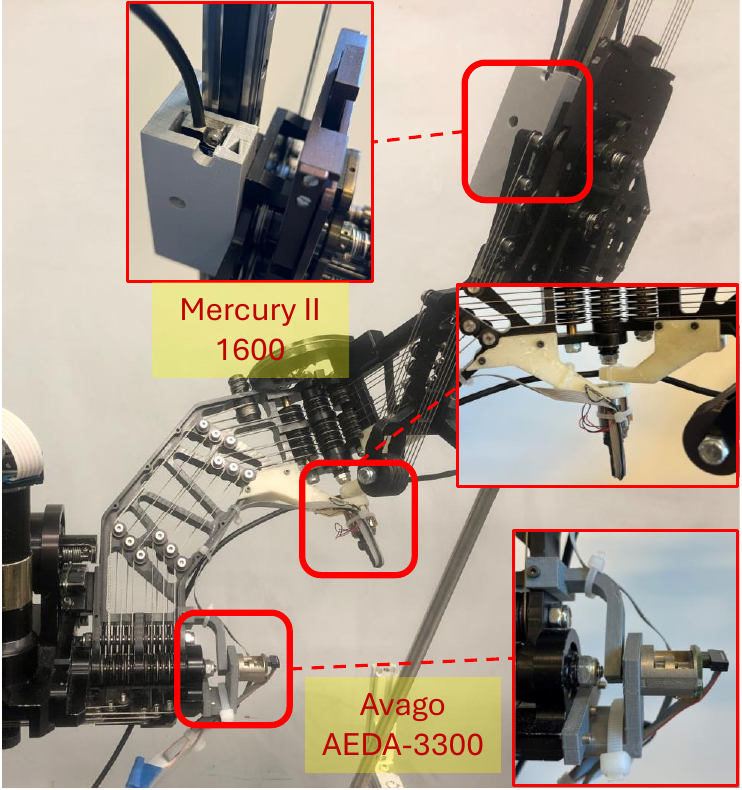}
\caption{The external joint encoders that are used to collect ground truth. After training, they can be removed in operation.} 
\label{fig_joint_encoder}
\end{figure}

The proposed methods were implemented and evaluated on a RAVEN-II surgical robot, which is fully cable-driven. Only the left arm of the robot was used. To collect reliable ground truth of joint positions, Avago Technologies AEDA-3300 encoders were installed on the rotational joints 1 and 2, with a resolution of 80000 PPR. Mercury II 1600 was installed on the translational joint 3, with a resolution of 5 {\textmu}m, as shown in Fig. \ref{fig_joint_encoder}. As described in \ref{sch_crtk_controller}, the external encoders were registered during the initialization of RAVEN-II, in which the limit of each joint was explored and reached. In this work, temporary joint encoders were used for the best accuracy and reliability of the ground truth. In real practice, for better feasibility, visual tracking might be preferred \cite{peng2020real, hwang2020efficiently, mahler2014learning}.

The hyperparameters of the DNN model were chosen empirically and are inspired by our previous work \cite{peng2020real, peng2023ablation} and the state-of-the-art work \cite{seita2018fast}. Unless stated otherwise, the DNN model for calibration in this paper has 2 fully connected hidden layers with 100 units each. The sigmoid function was used as activation for all hidden units. Mean squared error was used as the loss function. L2 kernel regularization was applied with a factor of 0.0005. The DNN model was trained by 200 epochs with a learning rate of 0.001 and batch size of 1024. Adam \cite{kingma2014adam} was used as the optimizer, and the exponential decay rates of the 1st and 2nd order moment estimates were 0.9 and 0.999, respectively. 

\subsection{Robot States of RAVEN-II} \label{schar_robot_states}

\begin{figure*}
\centering
\includegraphics[width=0.9\textwidth]{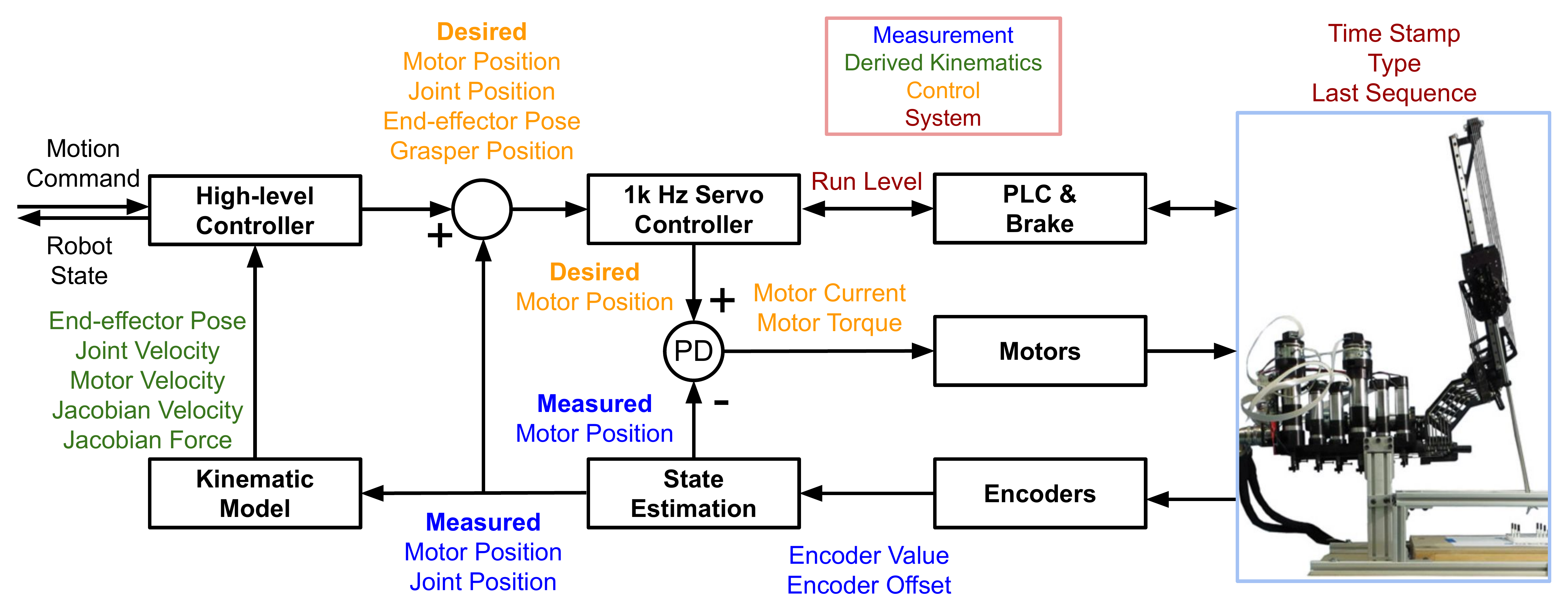}
\caption{The robot states and control system of the RAVEN-II surgical robot. Among the robot states, only encoder values are direct measurements from the encoders mounted on the motors on the robot base. The motor positions and joint positions are derived from encoder values and have a linear relationship. Motion commands can be given in either Cartesian space or joint space, and motor current commands are determined by PD control on measured and desired motor positions. The motor current in the robot state is the command instead of measurement, and the motor torque is derived from the motor current with a linear relationship.} 
\label{fig_ravenstate}
\end{figure*}

There are multiple original and derived measurements, control commands, and system statuses available in the robot state of RAVEN-II, which is published to a ROS topic 'ravenstate' at 1000 Hz. According to the ablation study \cite{peng2023ablation}, joint positions and motor torques are used as the input of the calibration models, unless stated otherwise. The robot states and control system of the RAVEN-II robot is shown in Fig. \ref{fig_ravenstate}.

\textbf{Operating status} includes the \textbf{time stamp}; \textbf{run level} -- a number indicating major run levels such as operating, pausing, and homing; \textbf{sublevel} -- a number indicating subordinate run levels, such as motor torque test; \textbf{last sequence} -- sequence number of teleoperation commands; arm \textbf{type} -- a number indicating left or right arm; \textbf{desired grasper position} -- a number indicating open or close of the grasper.

\textbf{Encoder values} are measurements from encoders on the motors mounted on the robot base. \textbf{Encoder offsets} are registered during the robot homing procedure when joint limits are explored. \textbf{Measured motor positions} are derived from encoder values and offsets. Similarly, \textbf{measured joint positions} are derived from the measured motor positions. Encoder values and offsets, measured motor positions, and measured joint positions have a linear relationship. Due to the cable-driven mechanism, the joint positions of RAVEN-II have considerable errors. Utilizing forward kinematics \cite{king2012kinematic}, \textbf{measured end-effector pose} is obtained by measured joint positions and has errors as well. By taking derivatives, \textbf{motor velocities} and \textbf{joint velocities} are obtained from previous and current motor positions and joint positions, respectively.

Control command of RAVEN-II can be given in joint space or Cartesian space, incrementally or absolutely. \textbf{Desired end-effector pose} is determined by the control command and the current end-effector pose. Next, \textbf{desired joint positions} are derived from the desired end-effector pose by inverse kinematics, which is further processed to \textbf{desired motor positions} with a linear relationship. RAVEN-II has closed-form inverse kinematics solutions reported in \cite{king2012kinematic}. Because the servo control loop of RAVEN-II runs on 1000 Hz, the difference between the current joint positions and the desired joint positions is minuscule. For example, a joint velocity of 10 deg/s results in a difference of 0.01 deg.

\textbf{Motor current command} is the DC current command determined by motor PD control based on the current and desired motor positions. And \textbf{motor torques} have a linear relationship with the motor current command. Thus, both states are control commands instead of measurements.

\textbf{Jacobian velocity \& force:} velocity and contact force of the end-effector, derived by joint velocities, motor torques, and the Jacobian matrix \cite{hannaford2023chapter5, craig1986introduction}. It is worth noticing that these states can be considerably inaccurate since the real robot is not purely kinematic with frictions, masses, and so on.

\section*{CODE AVAILABILITY}
The CRTK-Python controller for the RAVEN-II surgical robot is open-sourced at: \url{https://github.com/uw-biorobotics/raven2_CRTK_Python_controller}.

\bibliographystyle{IEEEtran}
\bibliography{IEEEabrv,IEEEexample}


\end{document}